\documentclass[sigconf]{acmart}
\usepackage{times}  
\usepackage{helvet}  
\usepackage{courier}  
\usepackage{url}  
\usepackage{graphicx}  

\usepackage{balance}
\usepackage{comment}
\usepackage{xcolor}
\usepackage{soul}
\usepackage[utf8]{inputenc}
\usepackage{amsfonts}
\usepackage{amsmath}
\usepackage{epsfig}
\usepackage{amsopn}
\usepackage{float}
\usepackage{epstopdf}

\usepackage{booktabs} 
\usepackage{algorithmic}
\usepackage{colortbl}
\usepackage[ruled]{algorithm2e} 

\usepackage[skip=2pt]{caption}  
\usepackage{subcaption}
\usepackage{multirow}
\usepackage{tabularx}
\usepackage{bm}
\usepackage{rotating}
\usepackage{mathrsfs}
\usepackage{scalerel}
\usepackage[flushleft]{threeparttable}
\usepackage{array}
\newcolumntype{L}[1]{>{\raggedright\let\newline\\\arraybackslash\hspace{0pt}}m{#1}}
\newcolumntype{C}[1]{>{\centering\let\newline\\\arraybackslash\hspace{0pt}}m{#1}}
\newcolumntype{R}[1]{>{\raggedleft\let\newline\\\arraybackslash\hspace{0pt}}m{#1}}

\begin{document}

\title{Adversarial Variational Embedding for Robust Semi-supervised Learning}

\author{Xiang Zhang, Lina Yao, Feng Yuan}
\affiliation{%
  \institution{University of New South Wales, Sydney, Australia}
}\email{xiang.zhang3@student.unsw.edu.au, lina.yao@unsw.edu.au, feng.yuan@student.unsw.edu.au}



\begin{abstract}
Semi-supervised learning is sought for leveraging the unlabelled data when labelled data is difficult or expensive to acquire. 
Deep generative models (e.g., Variational Autoencoder (VAE)) and semi-supervised Generative Adversarial Networks (GANs) have recently shown promising performance in semi-supervised classification for the excellent discriminative representing ability. However, the latent code learned by the traditional VAE is not exclusive (repeatable) for a specific input sample, which prevents it from excellent classification performance. In particular, the learned latent representation depends on a non-exclusive component which is stochastically sampled from the prior distribution. 
Moreover, the semi-supervised GAN models generate data from pre-defined distribution (e.g., Gaussian noises) which is independent of the input data distribution and may obstruct the convergence and is difficult to control the distribution of the generated data. 
To address the aforementioned issues, we propose a novel Adversarial Variational Embedding (AVAE) framework for robust and effective semi-supervised learning to leverage both the advantage of GAN as a high quality generative model and VAE as a posterior distribution learner. 
The proposed approach first produces an exclusive latent code by the model which we call VAE++, and meanwhile, provides a meaningful prior distribution for the generator of GAN. 
The proposed approach is evaluated over four different real-world applications and we show that our method outperforms the state-of-the-art models, which confirms that the combination of VAE++ and GAN can provide significant improvements in semi-supervised classification.
\end{abstract}
%
\copyrightyear{2019} 
\acmYear{2019} 
\setcopyright{acmcopyright}
\acmConference[KDD '19]{The 25th ACM SIGKDD Conference on Knowledge Discovery and Data Mining}{August 4--8, 2019}{Anchorage, AK, USA}
\acmBooktitle{The 25th ACM SIGKDD Conference on Knowledge Discovery and Data Mining (KDD '19), August 4--8, 2019, Anchorage, AK, USA}
\acmPrice{15.00}
\acmDOI{10.1145/3292500.3330966}
\acmISBN{978-1-4503-6201-6/19/08}

\keywords{Variational Autoencoder, Generative Adversarial Networks, Representation Learning, Semi-supervised Classification}

\maketitle

\section{Introduction} 
\label{sec:introduction}
Semi-supervised learning from data is one of the fundamental challenges in
artificial intelligence, which considers the problem when only a subset of the observations has corresponding class labels \cite{ghasedi2018semi}. This issue is of immense practical interest in a broad range of application scenarios, such as abnormal activity detection \cite{yao2016learning}, neurological diagnosis \cite{peng2016immune}, computer vision \cite{gong2016multi}, and recommender systems \cite{yang2017bridging}. In these scenarios, it is easy to obtain abundant observations but expensive to gather the corresponding class labels. Among existing approaches, Variational Autoencoders (VAEs) \cite{kingma2014semi,sonderby2016ladder} have recently achieved state-of-the-art performance in semi-supervised learning. 

VAE models provide a general framework for learning latent representations: a model is specified by a joint probability distribution both over the data and over latent random variables, and a representation can be found by considering the posterior on latent variables given specific data \cite{narayanaswamy2017learning}. The learned representations can not only be used for generation but also for classification. For instance, VAE provides a latent feature representation of the input observations, where a separate classifier can be thereafter trained using these representations. The high quality of latent representations enables accurate classification, even with a limited number of labels. 
 A number of studies have applied VAE in semi-supervised classification in the computer vision area \cite{kingma2014semi,makhzani2015adversarial,narayanaswamy2017learning}. 

\begin{figure}[t]
    \centering
    \begin{subfigure}[t]{0.23\textwidth}
        \centering
        \includegraphics[width=\textwidth]{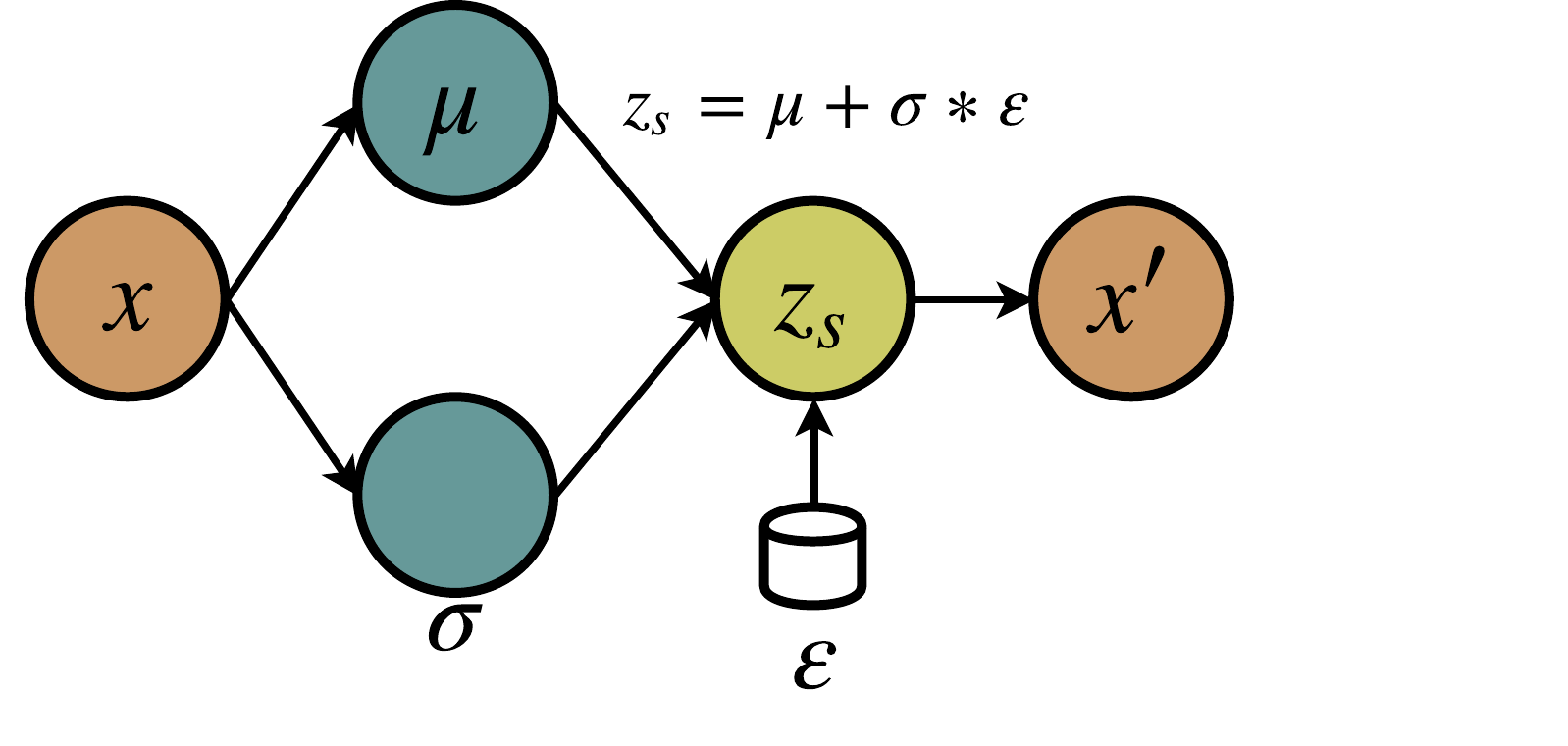}
        \caption{Standard VAE}
        \label{fig:standard_VAE}
    \end{subfigure}%
    \begin{subfigure}[t]{0.27\textwidth}
        \centering
        \includegraphics[width=\textwidth]{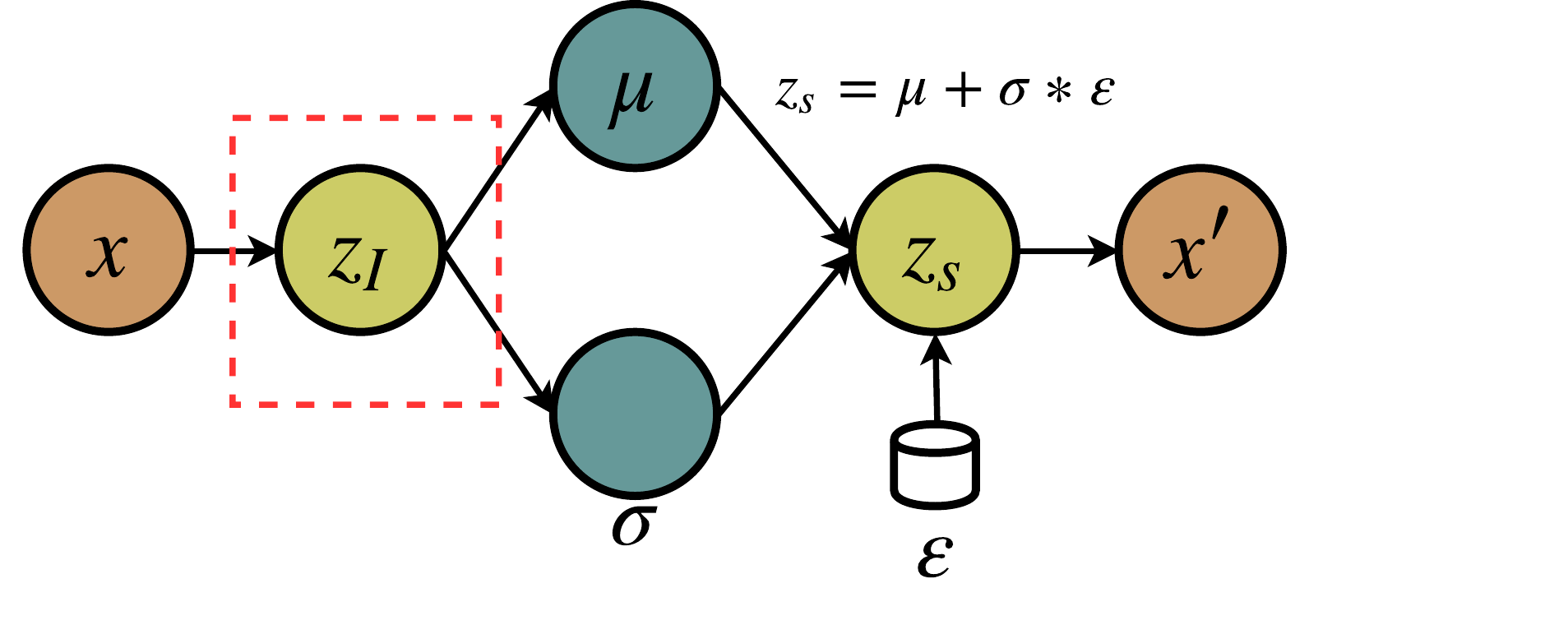}
        \caption{VAE++}
        \label{fig:inherited_GAN}
    \end{subfigure}%
    \caption{Comparison of the standard VAE and the proposed VAE++. $\bm{x}$ and $\bm{x'}$ denote the input and the reconstructed data. $\bm{\mu}$ and $\bm{\sigma}$ denote the learned expectation and standard deviation, $\bm{z}_s$ denotes the stochastically sampled latent representation which is composed by $\bm{\mu}$, $\bm{\sigma}$, and $\bm{\varepsilon}$, where $\bm{\varepsilon}$ is randomly sampled from $\mathcal{N}(0, 1)$. In standard VAE, $\bm{z}_s$ is regarded as the learned representation while, in VAE++, $\bm{z}_I$ denotes the proposed exclusive latent representation which can be used for classification. 
    }
    \label{fig:vae_structure}
    \vspace{-5mm}
\end{figure}

 \subsection{Motivation} 
\label{sub:motivation}

\textit{Why we propose the VAE++
.}  One major challenge faced by the existing VAE-based semi-supervised methods is that the latent representations are stochastically sampled from the prior distribution instead of being directly rendered from the explicit observations. 
In particular, as shown in Figure~\ref{fig:standard_VAE}, the learned latent representations $\bm{z}_s$ are \textit{randomly sampled} from a multivariate Gaussian distribution (see Equation~\ref{eq:eq1}). Thus, for a specific sample, the corresponding latent representation is not exclusive (i.e., the representation is not repeatable in different runnings), which makes it inappropriate for classification. To solve this problem, in the latent space, we propose a new variable $\bm{z}_I$ (see Figure~\ref{fig:inherited_GAN}) which is directly learned from the input data. The exclusive latent code $\bm{z}_I$ is guaranteed to keep invariant for a specific input $\bm{x}$ in different runnings. 
The modified VAE is called VAE++.
In addition, the learned expectation $\bm{\mu}$ only contains a part of information of the input observations, which is not enough to represent the observations in classification task, even though $\bm{\mu}$ is exclusive\footnote{For the same reason, $\bm{\sigma}$ can not be used as the exclusive code.}. The comparison of performance among $\bm{z}_I$, $\bm{z}_s$ and $\bm{\mu}$ will be presented in Section~\ref{sec:experiments}.

\textit{Why VAE++ needs the semi-supervised GAN.} In the proposed VAE++, it is necessary to reduce the information loss between the two latent representations $\bm{z}_I$ and $\bm{z}_s$ to guarantee the learned $\bm{z}_I$ is representative. The commonly used constraints between two distributions (e.g., Kullback-Leibler divergence) can only utilize the information of the observations but fail to exploit the information of labels. In this paper, we use a novel approach to take advantage of both unlabelled and labelled data by jointly training the VAE++ and a semi-supervised GAN. 

\textit{Why semi-supervised GAN needs the VAE++.} 
GAN based approaches \cite{odena2016semi,salimans2016improved} have recently shown promising results in semi-supervised learning. The semi-supervised GAN trains a generative model and a discriminator with inputs belonging to one of $K$ classes. Different from the regular GAN, the semi-supervised GAN requires the discriminator to make a $K+1$ class prediction with an extra class added, corresponding to the generated fake samples. In this way, the observations' properties can be used to improve decision boundaries and allow for more accurate classification than using the labelled data alone. 
However, the generated samples are sampled from pre-defined distribution (e.g., Gaussian noise) \cite{cao2018adversarial}. Such pre-defined prior distributions are often independent from the input data distributions and may obstruct the convergence and can not guarantee the distribution of the generated data. This drawback can be amended by gearing with VAE++ which can provide a meaningful prior distribution that can represent the distribution of the input data.


We introduce a recipe for semi-supervised learning, a robust Adversarial Variational Embedding (AVAE) framework, which learns the exclusive latent representations by combining VAE and semi-supervised GAN. To utilize the generative ability of GAN and the distribution approximating power of VAE, the proposed approach employs GAN to encourage VAE for the aim of learning the more robust and informative latent code.
We present the framework in the context of VAE, adding a new exclusive code in latent space which is directly rendered from the data space.
The generator in VAE++ also works as a generator of GAN.
Both the exclusive code (marked as real) and the generated representation (marked as fake) are fed into the discriminator in order to force them to have similar distribution
 \cite{mirza2014conditional}.

\subsection{Contribution} 
\label{sub:contribution}
Although a small set of models combining VAE and GAN have been previously explored, they are all focused on the generation perspective. To our knowledge, we are in the first batch of work that focuses on classification by aggregating VAE and GAN. We mark the following contributions:
\begin{itemize}
    \item We present a novel semi-supervised Adversarial Variational Embedding approach to harness the deep generative model and generative adversarial networks collectively under a trainable unified framework. The reproducible codes and datasets are publicly available\footnote{https://github.com/xiangzhang1015/Adversarial-Variational-Semi-supervised-Learning}.



    \item We propose a new structure, VAE++, to automatically learn an exclusive latent code for accurate classification.
    A novel semi-supervised GAN, which exploits both the unlabelled data distribution and categorical information, is proposed to gear with the VAE++ in order to
    encourage the VAE++ to learn a more effective and robust exclusive code.

    \item We evaluate the proposed approach over four real-world applications (activity reconstruction, neurological diagnosis, image classification, and recommender system). The results demonstrate that our approach outperforms all the state-of-the-art methods.

\end{itemize}
\vspace{-1mm}

\section{Related Work} 
\label{sec:related_work}


There are a host of studies that have been investigated to apply VAE for semi-supervised learning \cite{kingma2014semi,narayanaswamy2017learning,sonderby2016ladder,maaloe2016auxiliary}.
 \cite{kingma2014semi} explores semi-supervised learning with deep generative models by building two VAE-based deep generative models for latent representation extraction.
 Afterward, \cite{narayanaswamy2017learning} attempts to learn disentangled representations that encode distinct aspects of the data into separate variables. However, in all the existing semi-supervised VAE models, the learned representations do not only depend on the posterior distribution but also on the latent random variables. It is necessary that learning the exclusive code which is only related to the posterior distribution for the specified data.

Another recent arising semi-supervised method is semi-supervised GAN \cite{odena2016semi,springenberg2015unsupervised,radford2015unsupervised}. SGAN \cite{odena2016semi} extends GAN to the semi-supervised context by forcing the discriminator network to output class labels. The CatGAN \cite{springenberg2015unsupervised} modifies the objective function to take into account the mutual information between observed examples and their predicted class distributions. In the above methods, the generator chooses simple factored continuous noise which is independent from the input data distribution, for generation. As a result, it is possible that the noise will be used by the generator in a highly entangled way, increasing the difficulty to control the distribution of the generated data. Conditional GAN \cite{mirza2014conditional} and InfoGAN \cite{chen2016infogan} address this drawback by utilizing external information (e.g., categorical information) as a restriction, but they both pay attention to generation or supervised classification and have limited help in semi-supervised classification. 

Despite the few works attempting to combine VAE and GAN \cite{larsen2015autoencoding,makhzani2015adversarial,bao2017cvae}, most of them focus on generation instead of classification. For example, the VAE/GAN \cite{larsen2015autoencoding} and CVAE-GAN \cite{bao2017cvae} employ the standard VAE to share the encoder with the generator of GAN in order to generate new observations.
For semi-supervised classification, we care about the latent code instead of the observations. The Adversarial Autoencoder (AAE \cite{makhzani2015adversarial}) integrates VAE and GAN but only employs GAN to replace KL divergence as a penalty to impose a prior distribution on the latent code, which is a totally different direction from our work. 

\textbf{Summary.}  Unlike the existing VAE- and GAN-based studies, the proposed model 1) focuses on semi-supervised classification instead of generation; 2) attempts to learn an exclusive latent representation instead of a stochastic sampled representation; 3) works on improvement of latent space instead of data space. Moreover, the semi-supervised GAN in our work partly adopts the improved GAN \cite{salimans2016improved}, but there are a number of differences: 1) \cite{salimans2016improved} adopts the semi-supervised strategy for classification while we adopt this strategy as a constraint
to reduce information loss in the transformation from $z_I$ to $z_s$ in order to force the proposed AVAE to learn a more robust and effective latent code;  2) \cite{salimans2016improved} employs the discriminator of GAN as the classifier while we adopt an extra non-parametric classifier since the former has poor performance in our case 
(take the PAMAP2 dataset as an example, \cite{salimans2016improved} and our model achieve the accuracy around 65\% and 85\%, respectively); 3) we employ weighted loss function to balance the significance of the unlabelled and labelled observations.

\section{Methodology} 
\label{sec:methodology}
Suppose the input dataset has two subsets, one of which contains labelled samples while the other contains unlabelled samples. In the former subset, the observations appear as pairs $(\bm{X}^L, \bm{Y}^L) = \{(\bm{x}_1^L, \bm{y}_1), (\bm{x}_2^L, \bm{y}_2), \cdots, (\bm{x}_{N_L}^L, \bm{y}_{N_L})\}$ with the $i$-th observation $\bm{x}_i^L \in \mathbb{R}^M $ and the corresponding one-hot label $\bm{y}_i \in \mathbb{R}^K$ where $K$ denotes the number of classes. $N_L$ denotes the number of labelled observations while $M$ denotes the number of the observation dimensions. In the latter subset, only the observations $\bm{X}^U = \{\bm{x}_1^U, \bm{x}_2^U, \cdots, \bm{x}_{N_U}^U\}$ are available and $N_U$ denotes the number of unlabelled observations $\bm{x}_i^U \in \mathbb{R}^M $. The total data size $N$ equals to the sum of $N_L$ and $N_U$. In terms of effective classification, we attempt to learn a latent representation
which is rich of distinguishable information. Then the learned representations
can be fed into a classifier for recognition. In this paper, we mainly focus on the latent code learning. 

In the semi-supervised learning, due to the lack of labelled observations, it is significant to learn latent variable distribution
based on the observations without label\footnote{For simplification, we omit the index and directly use variable $\bm{x}$ to denote observations.}. Thus, we are required to build an encoder to provide an embedding or feature representation which allows accurate classification even with limited observations. 

\begin{figure}[t!]
\centering
\includegraphics[width=0.35\textwidth]{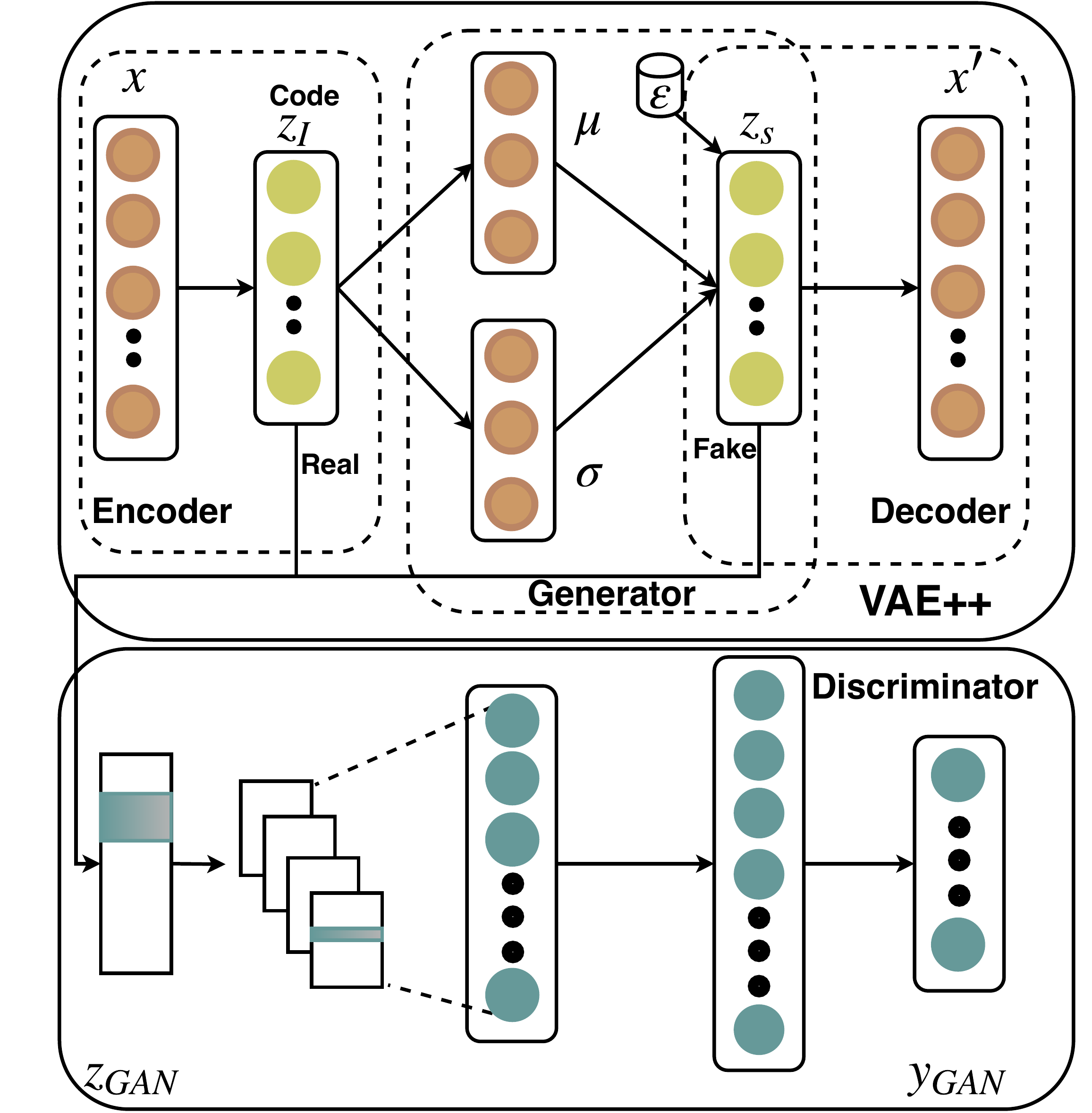}
\caption{AVAE is composed of VAE++ and a semi-supervised GAN. The generated $\bm{z}_s$ (labelled as fake) and the exclusive code $\bm{z}_I$ (labelled as real) are fed into the discriminator. The discriminator can exploit both the labelled and unlabelled observations. The generator in VAE++ also works as a generator of GAN.
}
\label{fig:AIVAE}
\vspace{-4mm}
\end{figure}

\subsection{VAE++} 
\label{sub:vae}
The VAE is demonstrated to provide a latent feature representation for semi-supervised learning \cite{kingma2014semi,narayanaswamy2017learning}, compared to a linear embedding method or a regular autoencoder. 
The VAE maps the input observation $\bm{x}$ to a compressed code $\bm{z}_s$, and decodes it to reconstruct the observation. The latent representation is calculated through the reparameterization trick \cite{kingma2013auto}:
\begin{equation} \label{eq:eq1}
\bm{z}_s = \mu_{\bm{x}} + \sigma_{\bm{x}}*\bm{\varepsilon}
\end{equation}
with $\bm{\varepsilon} \sim \mathcal{N}(0, 1)$ to impose the posterior distribution of the latent code on
$p(\bm{z}_s|x) \sim \mathcal{N}(\mu_{\bm{x}}, \sigma^2_{\bm{x}})$. $\mu_{\bm{x}}$ and $\sigma_{\bm{x}}$ denote the expectation and standard deviation of the posterior distribution of $\bm{z}_s$, which are learned from $\bm{x}$. 
For the efficient generation and reconstruction, VAE imposes the code $\bm{z}_s$ on a prior Gaussian distribution:
$$\bar{p}(\bm{z}_s) = \mathcal{N}(\bm{z}_s|\bm{0}, \bm{I})$$
Through minimizing the reconstruction error between $\bm{x}$ and $\bm{x}'$ and restricting the distribution of $\bm{z}_s$ to approximate the prior distribution $\bar{p}(\bm{z}_s)$, VAE is supposed to learn the representative latent code $\bm{z}_s$ which can be used for classification or generation.

Due to the strong feature representation ability, VAE has been employed for feature extraction and semi-supervised learning \cite{abbasnejad2017infinite,xu2017variational,walker2016uncertain,narayanaswamy2017learning}. However, one limitation of the standard VAE is that the learned latent code $\bm{z}_s =g(\mu_{\bm{x}}, \sigma_{\bm{x}}, \bm{\varepsilon})$, as shown in Equation~(\ref{eq:eq1}), is not exclusive. In other words, for a specific observation $\bm{x}$ and a fixed embedding model $p(\bm{z}_s|\bm{x})$, the corresponding latent code $\bm{z}_s$ is not exclusive as it contains a stochastic variable $\bm{\varepsilon}$ which is randomly sampled from the prior distribution $\bar{p}(\bm{z}_s)$. For instance, in a pre-trained fixed VAE encoder, the specific input $\bm{x}$ will lead to a variety of $\bm{z}_s$ in different running. 
At high level, the latent code $\bm{z}_s$ is determined by two factors: the prior distribution of observation $\bar{p}(\bm{x})$ which affects $\bm{z}_s$ through the learned $\mu_{\bm{x}}$ and $\sigma_{\bm{x}}$, and the stochastically sampled data $\bm{\varepsilon}$. 
However, the stochastically sampled latent code is unstable and will corrupt the features for classification. Furthermore, the posterior distribution of $\bm{z}_s$ is forced to approximate the manually set prior distribution (commonly Normal Gaussian distribution), which inevitably leads to information loss.

In order to completely sidestep the above-mentioned issue, in this paper, we propose a novel VAE++ model to learn an exclusive latent code $\bm{z}_I$. The VAE++ contains three key components: the encoder, the generator, and the decoder (see Figure~\ref{fig:AIVAE}). 
The encoder transforms the observation into a latent code $\bm{z}_I \in \mathbb{R}^D$ which is directly determined by the input $\bm{x}$. $D$ denotes the dimension of $\bm{z}_I$. 
We learn the: 
$$p_{\bm{\theta}_{en}}(\bm{z}_I|\bm{x}) = f(\bm{z}_I; \bm{x}, \bm{\theta}_{en})$$ 
where $f$ denotes a non-linear transformation while $\bm{\theta}_{en}$ denotes encoder parameters. The non-linear transformation $f$ is generally chosen as a deep neural network for the excellent ability of non-linear approximation.
Then, in the generator, we measure the expectation $\mu(\bm{z}_I)$ and the standard derivation $\sigma(\bm{z}_I)$ from the latent code $\bm{z}_I$
 and update Equation~(\ref{eq:eq1}). The generated variable $\bm{z}_s$ can be calculated by:
\begin{equation} \label{eq3}
\bm{z}_s = \mu(\bm{z}_I) + \sigma(\bm{z}_I)*\bm{\varepsilon}
\end{equation}
At last, the decoder is employed to reconstruct the sample:
$$p_{\bm{\theta}_{de}}(\bm{x}'|\bm{z}_s) = f'(\bm{x}'; \bm{z}_s, \bm{\theta}_{de})$$
where $f'$ denotes another non-linear rendering, called decoder, with parameters $\bm{\theta}_{de}$ and $\bm{x}'$ denotes the reconstructed observation. 

The loss function of VAE++ can be calculated by:

\begin{equation} \label{eq2}
\begin{split}
\mathcal{L}_{\scaleto{VAE}{3pt}} &= - \mathbb{E}_{\bm{z}_s\sim p_{\bm{\theta}_{en}}(\bm{z}_s|\bm{x})}[ \log p_{\bm{\theta}_{de}}(\bm{x}'|\bm{z}_s)] \\
 &+ KL(p_{\bm{\theta}_{en}}(\bm{z}_s|\bm{x})||\bar{p}(\bm{z}_s))
 \end{split}
\end{equation}
The first component is the reconstruction loss, which equals to the expected negative log-likelihood of the observation. This term encourages the decoder to reconstruct the observation $\bm{x}$ based on the sampling code $\bm{z}_s$ which is under Gaussian distribution. The lower reconstruction error indicates the encoder learned a better latent representation. The second component is the Kullback-Leibler divergence which measures the distance between the prior distribution of the latent code $\bar{p}(\bm{z}_s)$ and the posterior distribution $p(\bm{z}_s|\bm{x})$. This divergence reflects the information loss when we use $p(\bm{z}_s|\bm{x})$ to represent $\bar{p}(\bm{z}_s)$. 

In the latent space of the novel VAE++, there are two compressed informative codes $\bm{z}_I$ and $\bm{z}_s$. The former represents directly-encoded $\bm{x}$ whilst the latter is stochastically sampled from the posterior distribution
, which makes the former more suitable for classification. Therefore, we choose $\bm{z}_I$ as the compressed latent code in VAE++ instead of the $\bm{z}_s$ in standard VAE. 

From equation (\ref{eq3}), we can observe that the expectation and standard deviation of $\bm{z}_s$ and $\bm{z}_I$ are invariant. In particular, for a specific sample $\bm{x}_i$, the corresponding $\bm{z}_{si}$ and $\bm{z}_{Ii}$ have the same statistical characteristics. Thus, we have $$\bm{z}_s \leftarrow \mu(\bm{z}_I), \sigma(\bm{z}_I), \bm{\varepsilon}$$ which indicates that the generated $\bm{z}_s$ is affected by both the distribution
  (or statistic characteristics) 
 of $\bm{z}_I$ and the prior distribution $\bar{p}(\bm{z}_s)$ (or $\bm{\varepsilon}$). In summary, the $\bm{z}_s$ inherits the statistical characteristics of $\bm{z}_I$.


\subsection{Adversarial Variational Embedding} 
\label{sub:svae_gan}
One significant sufficient condition of a well-trained VAE++ is less information loss in the transformation from $\bm{z}_I$ to $\bm{z}_s$ to guarantee the learned $\bm{z}_I$ is representative. As mentioned before, the information in $\bm{z}_s$ is partly inherited from $\bm{z}_I$ and the other part is randomly sampled from the prior distribution $\bar{p}(\bm{z}_s)$. Since the conditional distribution $p_{\bm{\theta}_{en}}(\bm{z}_I|\bm{x})$ has a better description of the input observation $\bm{x}$, we attempt to increase the proportion of inherited part
 and decrease the proportion of stochastically sampled part.

As shown in Figure~\ref{fig:AIVAE}, in the proposed AVAE the generator $\bm{G}$ generates $\bm{z}_s$ based on the joint probability $p(\mu, \sigma, \bar{p}(\bm{z}_s))$ instead of the noise in standard GAN. 
The $\bm{z}_s$ is regarded as `fake' while $\bm{z}_I$ is marked as `real'. 
Specifically, for the labelled observations $\bm{x}^L$, VAE++ encodes the input to the latent code $\bm{z}^L_I \in \mathbb{R}^D$ and generates $\bm{z}^L_s \in \mathbb{R}^D$; similarly, for unlabelled observations $\bm{x}^U$, we have $\bm{z}^U_I \in \mathbb{R}^D$ and generates $\bm{z}^U_s \in \mathbb{R}^D$. To exploit the information of the labels, we extend the $\bm{y} \in \mathbb{R}^K$ which has $K$ possible classes to $\bm{y}_{\scaleto{GAN}{3pt}} \in \mathbb{R}^{K+1}$ which has $K+1$ possible classes by regarding the generated fake samples $\bm{z}_s$ as the $(K+1)$-th class \cite{salimans2016improved,odena2016semi}. In the VAE++, the unspecified $\bm{z}_s$ denotes both $\bm{z}^L_s$ and $\bm{z}^U_s$ whenever we don't care whether the observation is labelled or not. This rule also applies to $\bm{z}_I$. Similarly, we use $\bm{z}_{\scaleto{GAN}{3pt}}$ to denote the input of the discriminator $\bm{D}$, which contains both $\bm{z}_I$ and $\bm{z}_s$. The discriminator can be described by 
$$\bm{q}_{\bm{\varphi}}(\bm{y}_{\scaleto{GAN}{3pt}}|\bm{z}_{\scaleto{GAN}{3pt}}) = h(\bm{y}_{\scaleto{GAN}{3pt}}; \bm{z}_{\scaleto{GAN}{3pt}},\bm{\varphi})$$
where $\bm{\varphi}$ denotes the parameters of $\bm{D}$ while $h$ denotes the non-linear transformation which is implemented by a Convolutional Neural Networks (CNN) \cite{krizhevsky2012imagenet} in this paper. Therefore, we can use $\bm{q}_{\bm{\varphi}}(\bm{y}_{\scaleto{GAN}{3pt}}=K+1|\bm{z}_{\scaleto{GAN}{3pt}})$ to supply the probability where $\bm{z}_{\scaleto{GAN}{3pt}}$ is fake (from $\bm{z}_s$) and use 
$\bm{q}_{\bm{\varphi}}(\bm{y}_{\scaleto{GAN}{3pt}}|\bm{z}_{\scaleto{GAN}{3pt}}, \bm{y}_{\scaleto{GAN}{3pt}}<K+1)$ to supply the probability where $\bm{z}_{\scaleto{GAN}{3pt}}$ is real ((from $\bm{z}_I$)) and is correctly classified. 

For the labelled input, same as supervised learning, the discriminator is supposed to not only tell whether the input $\bm{z}_{\scaleto{GAN}{3pt}}$ is real or generated, but also classify it into the correct class. Therefore, we have the supervised loss function
$$\mathcal{L}_{\scaleto{label}{5pt}} = - \mathbb{E}_{\bm{z}_{\scaleto{GAN}{3pt}}, \bm{y}_{\scaleto{GAN}{3pt}}\sim p_{j}}[log \bm{q}_{\bm{\varphi}}(\bm{y}_{\scaleto{GAN}{3pt}}|\bm{z}_{\scaleto{GAN}{3pt}}, \bm{y}_{\scaleto{GAN}{3pt}}<K+1)]$$
where $p_j$ denotes the joint probability.

For the unlabelled input, we only require the discriminator to perform a binary classification: the input is real or fake. The former probability can be calculated by $1 - \bm{q}_{\bm{\varphi}}(\bm{y}_{\scaleto{GAN}{3pt}}=K+1|\bm{z}_{\scaleto{GAN}{3pt}})$ whilst the latter can be calculated by $\bm{q}_{\bm{\varphi}}(\bm{y}_{\scaleto{GAN}{3pt}}=K+1|\bm{z}_{\scaleto{GAN}{3pt}})$. Thus, the unsupervised loss function:
\begin{align*}
&\mathcal{L}_{\scaleto{unlabel}{5pt}}=   
 - \mathbb{E}_{\bm{z}_{\scaleto{GAN}{3pt}}\sim p_{\bm{\theta}_{en}}(\bm{z}_I|\bm{x})}[log(1 - \bm{q}_{\bm{\varphi}}(\bm{y}_{\scaleto{GAN}{3pt}}=K+1|\bm{z}_{\scaleto{GAN}{3pt}}))]\\ 
& \indent \indent \indent \indent - \mathbb{E}_{\bm{z}_{\scaleto{GAN}{3pt}}\sim p_{\bm{\theta}_{en}}(\bm{z}_s|\bm{x})}[log(\bm{q}_{\bm{\varphi}}(\bm{y}_{\scaleto{GAN}{3pt}}=K+1|\bm{z}_{\scaleto{GAN}{3pt}}))]
\end{align*}

In summary, the final loss function of the discriminator
\begin{equation} \label{eq4}
    \mathcal{L}_{\scaleto{GAN}{3pt}} = w_1*flag* \mathcal{L}_{\scaleto{label}{5pt}} + w_2*(1-flag) *\mathcal{L}_{\scaleto{unlabel}{5pt}}
\end{equation}
where  $w_1, w_2$ are weights and $flag$ is a switch function
$$flag = \left\{\begin{matrix}
1 & labelled\\ 
0 & unlabelled
\end{matrix}\right.$$
If the specific observation is labelled, we calculate the labelled loss function. Otherwise, we calculate the unlabelled loss function. From empirical experiments, we observe that the $\mathcal{L}_{\scaleto{unlabel}{5pt}}$ is much easier to converge than $\mathcal{L}_{\scaleto{label}{5pt}}$ and the real/fake classification accuracy is much higher than the $K$ classes classification accuracy. To encourage the optimizer to focus on the former part which is more difficult to converge, we set $w_1 = 0.9$ and $w_2 = 0.1$.

The discriminator receives $\bm{z}_{\scaleto{GAN}{3pt}}$ as input and extracts the dependencies through CNN filters. Two fully connected layers follow the convolutional layer for dimension reduction. At last, a softmax layer is employed to work on the low-dimension features to estimate the log normalization of the categorical probability distribution which is output as $\bm{y}_{\scaleto{GAN}{3pt}}$. 

The overall aim of the proposed AVAE (as described in Algorithm~\ref{alg:algorithm}) is to train a robust and effective semi-supervised embedding method. The VAE loss $\mathcal{L}_{\scaleto{VAE}{3pt}}$ and the GAN loss $\mathcal{L}_{\scaleto{GAN}{3pt}}$ are trained simultaneously by the Adam optimizer. After convergence, the compressed representative code $\bm{z}_I$ is fed into a non-parametric nearest neighbors classifier for recognition. 

\begin{algorithm}[!t]
\begin{small}
\caption{Adversarial Variational Embedding}
\label{alg:algorithm}
\begin{algorithmic}[1]
\renewcommand{\algorithmicrequire}{\textbf{Input:}}
\renewcommand{\algorithmicensure}{\textbf{Output:}}
 \REQUIRE labelled observations ($\bm{X}^L$, $\bm{Y}^L$) and unlabelled observations $\bm{X}^U$
 \ENSURE  Representation $\bm{z}_I$ 
 \STATE Initialize network parameters $\bm{\theta}_{en}$, $\bm{\theta}_{de}$, $\bm{q}_{\bm{\varphi}}$
 \FOR{$\bm{x} \in \{\bm{X}^L, \bm{X}^U\}$} 
    \STATE $ \bm{z}_I \gets \bm{x}$ 
    \STATE $ \bm{\mu}, \bm{\sigma} \gets \bm{z}_I$ 
    \STATE Sampling $\bm{\varepsilon}$ from $\mathcal{N}(\bm{0}, \bm{I})$ 
    \STATE $\bm{z}_s = \mu(\bm{z}_I) + \sigma(\bm{z}_I)*\bm{\varepsilon}$
    \STATE $\bm{x}' \gets \bm{z}_s$
    \STATE $\mathcal{L}_{\scaleto{VAE}{3pt}} \gets \bm{x}, \bm{x}', p(\bm{z}_s|\bm{x})$
    \FOR{${z}_I, {z}_s, \bm{y} \in \bm{Y}^L$}
        \STATE $\bm{y}_{\scaleto{GAN}{3pt}} \gets {z}_I, {z}_s$ 
        \STATE $\mathcal{L}_{\scaleto{GAN}{3pt}} \gets \bm{y}_{\scaleto{GAN}{3pt}}, \bm{y}$
    \ENDFOR
    \STATE Minimize $\mathcal{L}_{\scaleto{VAE}{3pt}}$ and $\mathcal{L}_{\scaleto{GAN}{3pt}}$
\ENDFOR
\RETURN $\bm{z}_I$
\end{algorithmic}
\vspace{-1mm}
\end{small}
\end{algorithm}

\begin{table*}[]
\centering
\caption{Overall comparison of semi-supervised classification accuracy (\%) on activity recognition. All the baselines and our approach are working on the same dataset and sharing the same experiment settings for each specific application.}
\label{tab:results_pamap2}
  \begin{threeparttable}
  \resizebox{\linewidth}{!}{
\begin{tabular}{C{2.2cm}|c|cccc|cccc|ccc|c}
\hline
\multirow{2}{*}{\textbf{Dataset}} & \multirow{2}{*}{\textbf{Rate (\%)}} & \multicolumn{4}{c|}{\textbf{Algorithm-related State-of-the-art}} & \multicolumn{4}{c|}{\textbf{Application-related State-of-the-art}} & \multicolumn{3}{c|}{\textbf{Ablation Study}} & \textbf{Ours} \\
 &  & \textbf{M2} & \textbf{AAE} & \textbf{LVAE} & \textbf{ADGM} & \textbf{\cite{chen2018interpretable}} & \textbf{\cite{lara2012centinela}} & \textbf{\cite{guo2016wearable}} & \textbf{\cite{zhang2018multi}} & \textbf{VAE ($\bm{\mu}$)} & \textbf{VAE} & \textbf{VAE++} & \textbf{AVAE} \\ \hline
\multirow{5}{*}{\textbf{\begin{tabular}[c]{@{}l@{}}Activity\\ Recognition\\ (PAMAP2)\end{tabular}}} & 20 & 64.83$\pm$0.16 & 63.67$\pm$0.23 & 69.82$\pm$0.69 & 67.31$\pm$0.45 & 72.31$\pm$0.16 & 70.95$\pm$0.08 & 67.31$\pm$0.14 & 76.68$\pm$0.31 & 58.43$\pm$0.13 & 76.51$\pm$0.53 & 78.12$\pm$0.55 & 78.63$\pm$0.38 \\
 & 40 & 68.92$\pm$0.23 & 76.83$\pm$0.25 & 76.43$\pm$0.19 & 78.21$\pm$0.38 & 80.51$\pm$0.21 & 75.38$\pm$0.12 & 77.28$\pm$0.21 & 80.15$\pm$0.16 & 62.74$\pm$0.12 & 78.78$\pm$0.22 & 80.88$\pm$0.38 & 81.37$\pm$0.29 \\
 & 60 & 72.35$\pm$0.21 & 77.39$\pm$0.19 & 78.69$\pm$0.27 & 79.34$\pm$0.29 & 80.29$\pm$0.21 & 76.89$\pm$0.05 & 79.69$\pm$0.15 & 82.49$\pm$0.33 & 67.85$\pm$0.08 & 79.63$\pm$0.29 & 81.94$\pm$0.19 & 84.91$\pm$0.17 \\
 & 80 & 75.88$\pm$0.35 & 78.28$\pm$0.11 & 81.41$\pm$0.23 & 80.38$\pm$0.16 & 82.12$\pm$0.16 & 79.95$\pm$0.18 & 81.65$\pm$0.09 & 83.56$\pm$0.11 & 73.43$\pm$0.06 & 81.75$\pm$0.17 & 82.08$\pm$0.26 & 85.56$\pm$0.21 \\
 & 100 & 77.59$\pm$0.17 & 80.79$\pm$0.14 & 84.39$\pm$0.18 & 83.66$\pm$0.16 & 83.64$\pm$0.12 & 81.96$\pm$0.11 & 82.38$\pm$0.13 & 84.59$\pm$0.24 & 76.85$\pm$0.00 & 82.37$\pm$0.25 & 83.29$\pm$0.18 & 86.41$\pm$0.06 \\ \hline
\end{tabular}
}
\begin{tablenotes}
  \small
  \item Note: If the compared method can not deal with unsupervised samples, it will be trained only by the supervised samples.
\end{tablenotes}
\end{threeparttable}
\end{table*}

\begin{table*}[]
\centering
\caption{Overall comparison of semi-supervised classification accuracy (\%) on neurological diagnosis}
\label{tab:results_tuh}
 \resizebox{\linewidth}{!}{
\begin{tabular}{C{2.2cm}|c|cccc|cccc|ccc|c}
\hline
\multirow{2}{*}{\textbf{Dataset}} & \multirow{2}{*}{\textbf{Rate (\%)}} & \multicolumn{4}{c|}{\textbf{Algorithm-related State-of-the-art}} & \multicolumn{4}{c|}{\textbf{Application-related State-of-the-art}} & \multicolumn{3}{c|}{\textbf{Ablation Study}} & \textbf{Ours} \\
 &  & \textbf{M2} & \textbf{AAE} & \textbf{LVAE} & \textbf{ADGM} &  \textbf{\cite{ziyabari2017objective}} & \textbf{\cite{harati2015improved}} & \textbf{\cite{schirrmeister2017deep}} & \textbf{\cite{goodwin2017deep}} & \textbf{VAE ($\bm{\mu}$)} & \textbf{VAE} & \textbf{VAE++} & \textbf{AVAE} \\ \hline
\multirow{5}{*}{\textbf{\begin{tabular}[c]{@{}l@{}}Neurological \\ Diagnosis\\ (TUH)\end{tabular}}}& 20 & 71.28$\pm$0.16 & 80.13$\pm$0.95 & 82.31$\pm$0.19 & 86.32$\pm$0.12 & 87.66$\pm$0.23 & 86.38$\pm$0.36 & 82.19$\pm$0.24 & 86.33$\pm$0.21 & 80.58$\pm$0.69 & 86.37$\pm$0.24 & 0.86$\pm$0.53 & 93.69$\pm$0.16 \\
 & 40 & 75.32$\pm$0.16 & 82.95$\pm$0.26 & 84.38$\pm$0.16 & 86.99$\pm$0.05 & 89.25$\pm$0.19 & 91.58$\pm$0.35 & 84.21$\pm$0.08 & 89.25$\pm$0.34 & 81.35$\pm$0.24 & 89.69$\pm$0.27 & 91.28$\pm$0.25 & 94.32$\pm$0.28 \\
 & 60 & 76.32$\pm$0.29 & 86.21$\pm$0.52 & 87.51$\pm$0.26 & 87.65$\pm$0.16 & 91.28$\pm$0.37 & 92.58$\pm$0.26 & 85.36$\pm$0.32 & 90.38$\pm$0.24 & 82.59$\pm$0.63 & 90.58$\pm$0.27 & 92.87$\pm$0.31 & 95.21$\pm$0.21 \\
 & 80 & 79.65$\pm$0.37 & 88.53$\pm$0.28 & 89.56$\pm$0.25 & 88.05$\pm$0.12 & 92.59$\pm$0.26 & 93.25$\pm$0.31 & 85.16$\pm$0.24 & 91.59$\pm$0.16 & 83.21$\pm$0.21 & 91.69$\pm$0.35 & 93.96$\pm$0.28 & 97.86$\pm$0.26 \\
 & 100 & 82.59$\pm$0.31 & 89.58$\pm$0.25 & 90.25$\pm$0.21 & 88.65$\pm$0.26 & 93.32$\pm$0.18 & 94.29$\pm$0.25 & 86.42$\pm$0.26 & 92.4$\pm$0.25 & 84.21$\pm$0.65 & 92.38$\pm$0.41 & 94.65$\pm$0.24 & 98.13$\pm$0.32\\ \hline
\end{tabular}
}
\end{table*}

\begin{table*}[]
\caption{Overall comparison of semi-supervised classification accuracy (\%) on image classification}
\label{tab:results_mnist}
 \resizebox{\linewidth}{!}{
\begin{tabular}{C{2.2cm}|c|cccc|cccc|ccc|c}
\hline
\multirow{2}{*}{\textbf{Dataset}} & \multirow{2}{*}{\textbf{Rate (\%)}} & \multicolumn{4}{c|}{\textbf{Algorithm-related State-of-the-art}} & \multicolumn{4}{c|}{\textbf{Application-related State-of-the-art}} & \multicolumn{3}{c|}{\textbf{Ablation Study}} & \textbf{Ours} \\
 &  & \textbf{M2} & \textbf{AAE} & \textbf{LVAE} & \textbf{ADGM} & \textbf{\cite{odena2016semi}} & \textbf{\cite{springenberg2015unsupervised}} & \textbf{\cite{weston2012deep}} & \textbf{\cite{miyato2018virtual}} & \textbf{VAE ($\bm{\mu}$)} & \textbf{VAE} & \textbf{VAE++} & \textbf{AVAE} \\ \hline
\multirow{5}{*}{\textbf{\begin{tabular}[c]{@{}l@{}}Image\\ Classification\\ (MNIST)\end{tabular}}} & 20& 93.22$\pm$0.62 & 90.25$\pm$0.25 & 93.25$\pm$0.26 & 89.61$\pm$0.27 & 95.23$\pm$0.34 & 94.25$\pm$0.13 & 94.58$\pm$0.25 & 92.96$\pm$0.28 & 91.58$\pm$0.24 & 92.31$\pm$0.53 & 93.59$\pm$0.31 & 95.12$\pm$0.19 \\
 & 40 & 93.25$\pm$0.34 & 93.21$\pm$0.23 & 93.28$\pm$0.46 & 91.58$\pm$0.25 & 95.27$\pm$0.53 & 95.56$\pm$0.08 & 95.21$\pm$0.26 & 93.21$\pm$0.56 & 93.65$\pm$0.21 & 94.21$\pm$0.19 & 94.68$\pm$0.28 & 96.43$\pm$0.35 \\
 & 60 & 96.24$\pm$0.51 & 96.35$\pm$0.27 & 95.34$\pm$0.21 & 93.21$\pm$0.34 & 96.38$\pm$0.22 & 96.54$\pm$0.08 & 96.48$\pm$0.32 & 96.28$\pm$0.57 & 94.89$\pm$0.21 & 95.34$\pm$0.14 & 96.42$\pm$0.25 & 97.21$\pm$0.21 \\
 & 80 & 98.19$\pm$0.25 & 95.32$\pm$0.37 & 96.11$\pm$0.52 & 95.01$\pm$0.15 & 97.82$\pm$0.11 & 97.21$\pm$0.13 & 97.86$\pm$0.34 & 97.63$\pm$0.15 & 96.78$\pm$0.25 & 97.63$\pm$0.15 & 98.71$\pm$0.16 & 99.79$\pm$0.12 \\
 & 100 & 98.65$\pm$0.21 & 0.98.25$\pm$0.61 & 96.35$\pm$0.26 & 95.38$\pm$0.82 & 99.21$\pm$0.26 & 98.64$\pm$0.27 & 99.06$\pm$0.22 & 98.53$\pm$0.17 & 97.41$\pm$0.18 & 98.35$\pm$0.09 & 99.67$\pm$0.23 & 99.85$\pm$0.11  \\ \hline
\end{tabular}
}
\end{table*}

\begin{table*}[]
\caption{Overall comparison of semi-supervised classification accuracy (\%) on recommender system}
\label{tab:results_yelp}
 \resizebox{\linewidth}{!}{
\begin{tabular}{C{2.2cm}|c|cccc|cccc|ccc|c}
\hline
\multirow{2}{*}{\textbf{Dataset}} & \multirow{2}{*}{\textbf{Rate (\%)}} & \multicolumn{4}{c|}{\textbf{Algorithm-related State-of-the-art}} & \multicolumn{4}{c|}{\textbf{Application-related State-of-the-art}} & \multicolumn{3}{c|}{\textbf{Ablation Study}} & \textbf{Ours} \\
 &  & \textbf{M2} & \textbf{AAE} & \textbf{LVAE} & \textbf{ADGM} & \textbf{\cite{pazzani-2007-nn}} & \textbf{\cite{rendle-2012-libfm}} & \textbf{\cite{he-2017-nfm}} & \textbf{\cite{chen-2017-acf}} & \textbf{VAE ($\bm{\mu}$)} & \textbf{VAE} & \textbf{VAE++} & \textbf{AVAE} \\ \hline
\multirow{5}{*}{\textbf{\begin{tabular}[c]{@{}l@{}}Recommender\\ System\\ (Yelp)\end{tabular}}} &  & 66.42$\pm$0.17 & 58.27$\pm$0.35 & 66.35$\pm$0.36 & 54.27$\pm$0.38 & 40.55$\pm$0.27 & 47.58$\pm$0.36 & 65.99$\pm$0.62 & 66.21$\pm$0.24 & 64.28$\pm$0.12 & 64.39$\pm$0.62 & 65.58$\pm$0.37 & 70.19$\pm$0.87 \\
 & 20 & 69.36$\pm$0.37 & 61.55$\pm$0.62 & 68.16$\pm$0.24 & 55.35$\pm$0.26 & 40.28$\pm$0.32 & 48.65$\pm$0.27 & 67.53$\pm$0.31 & 66.59$\pm$0.29 & 64.37$\pm$0.25 & 67.23$\pm$0.95 & 71.05$\pm$0.29 & 72.21$\pm$0.35 \\
 & 40 & 72.58$\pm$0.19 & 62.15$\pm$0.39 & 68.59$\pm$0.93 & 57.63$\pm$0.23 & 42.15$\pm$0.16 & 50.95$\pm$0.24 & 66.58$\pm$0.29 & 67.95$\pm$0.38 & 67.56$\pm$0.35 & 69.58$\pm$0.37 & 72.19$\pm$0.62 & 75.34$\pm$0.35 \\
 & 60 & 72.39$\pm$0.64 & 62.89$\pm$0.62 & 74.28$\pm$0.37 & 58.34$\pm$0.15 & 43.21$\pm$0.15 & 52.15$\pm$0.38 & 67.65$\pm$0.31 & 68.23$\pm$0.15 & 69.25$\pm$0.18 & 71.39$\pm$0.56 & 73.21$\pm$0.58 & 78.54$\pm$0.38 \\
 & 80 & 74.58$\pm$0.62 & 63.51$\pm$0.86 & 72.59$\pm$0.36 & 59.58$\pm$0.23 & 45.86$\pm$0.22 & 54.10$\pm$0.12 & 68.03$\pm$0.17 & 70.61$\pm$0.25 & 73.24$\pm$0.68 & 73.28$\pm$0.69 & 76.53$\pm$0.28 & 79.38$\pm$0.59\\ \hline
\end{tabular}
}
\end{table*}

\section{Experiments} 
\label{sec:experiments}
In this section, we demonstrate the effectiveness and validation of the proposed method over four applications.

\subsection{Activity Recognition} 
\label{sub:activity_recognition}
\subsubsection{Experiment Setup} 
\label{sub:experiment_setup_1}
Activity recognition is an important area in data mining.
We evaluate our approach over the well-known PAMAP2 dataset \cite{fida2015real}, which is collected by 9 participants (8 males and 1 female) aged $27\pm 3$.
We select 5 most commonly used activities (Cycling, standing, walking, lying, and running, labelled from 0 to 4)
as a subset for evaluation. For each subject, there are 12,000 instances. The activity is measured by 3 Inertial Measurement Units (IMU) attached to the participants' wrist, chest, and the outer ankle. Each IMU includes 13 dimensions: two 3-axis accelerometers, one 3-axis gyroscopes, one 3-axis magnetometers and one thermometer.
The experiments are performed by a Leave-One-Subject-Out strategy to ensure the practicality. 

The time window is set as 10 
with 50\% overlapping. The dataset is split into a training set (80\% proportion) and a testing set (20\% proportion). 
For semi-supervised learning, the training dataset contains both labelled observations and unlabelled observations. We present a term called`supervision rate' as a handle on the relative weight between the supervised and unsupervised terms. For the given number of labelled observations $N^L$ and the number of unlabelled observations $N^U$, the supervision rate $\gamma$ is defined by $N^L/(N^L+N^U)$. 

\subsubsection{Parameter Setting} 
\label{sub:parameter setting_1}
We introduce the default parameter settings and the settings in other applications keep the same if not mentioned.
The input observations are first normalized by Z-score normalization and fed to the input layer of the unsupervised VAE++. 
The neuron amount in the first hidden layer, which is denoted by $\bm{z}_I$, is a quarter of $M$. The second hidden layer contains 2 components which represent the expectation and the standard deviation respectively.
The third hidden layer $\bm{z}_s$ has the sample shape with $\bm{z}_I$. 
An Adam optimizer with a learning rate of $0.00001$ is employed to minimize the loss function of VAE++. 

After each epoch of VAE++, the first hidden layer $\bm{z}_I$ and the third hidden layer $\bm{z}_s$ are labelled as `real' and `fake', respectively, and fed to the discriminator $\mathcal{D}$. The discriminator contains one convolutional layer followed by two fully-connected layers. There is a softmax layer to obtain the categorical probability before the output layer which has $K+1$ neurons.
 The convolutional layer has 10 filters which have shape $[2, 2]$ and the stride size $[1,1]$. The padding method of the convolutional operation is set as `same' while the activation function is ReLU. The following hidden layer has $M/4$ neurons and the sigmoid activation function. The loss function is optimized by Adam update rule with learning rate of $0.0001$. The object functions of the VAE++ and the discriminator are trained simultaneously. After the convergence of the proposed method, the semi-supervised learned latent representation $\bm{z}_I$ is fed into a supervised non-parametric nearest neighbor classifiers with $k = 3$. 

 \subsubsection{Baselines} 
 \label{sub:baselines}
To measure the effectiveness of the proposed method, we compare it with a set of competitive state-of-the-art models. The state-of-the-art methods are composed of two categories: algorithm-related and application-related. The former denotes other VAE/GAN based semi-supervised classification algorithms, which are the same for all the applications. The comparison is used to demonstrate our framework has the highest semi-supervised representation learning ability. The latter denotes the state-of-the-art models in each application, which are varied for the different applications. The comparison is used to demonstrate our work is effective in the real-world scenarios.

The algorithm-related semi-supervised learning solutions in our comparison are listed as follows:
\begin{itemize}
    \item M2. \cite{kingma2014semi} proposes a probabilistic model that describes the data as being generated by a latent class variable in addition to a continuous latent representation.
    \item Adversarial Autoencoders (AAE). \cite{makhzani2015adversarial} employs the GAN to perform variational inference by matching the aggregated posterior of the hidden representation of the autoencoder.
    \item Ladder Variational Autoencoders (LVAE). \cite{sonderby2016ladder} proposes an inference model which recursively corrects the generative distribution by a data dependent likelihood.
    \item Auxiliary Deep Generative Models (ADGM). \cite{maaloe2016auxiliary} extends deep generative models with auxiliary variables, which improves the variational approximation.
\end{itemize}

We design ablation study to demonstrate the necessity of each key component of the proposed approach. In the ablation study, we set four control experiments with single variable among the components of AVAE. We adopt the following four methods to discover the latent representations: 1) VAE ($\bm{\mu}$) with $\bm{\mu}$ as the latent representation; 2) 
standard VAE ($\bm{z}_s$ as the latent representation); 3) 
VAE++ ( $\bm{z}_I$ as the latent representation); 4) 
AVAE. The extracted representations are fed into the same classifier for final classification. 

The application-related state-of-the-art models on activity recognition are listed here:
\begin{itemize}
	\item Chen et al. \cite{chen2018interpretable} adopt an attention mechanism to select the most distinguishable features from the activity signals and send them to a CNN structure for classification.
	\item Lara et al. \cite{lara2012centinela} apply both statistical and structural detectors features 
	to discriminate among activities.
	\item Guo et al. \cite{guo2016wearable} exploit the diversity of base classifiers to construct a good ensemble for multimodal activity recognition, and the diversity measure is obtained from both labelled and unlabelled data.
	\item Zhang et al. \cite{zhang2018multi} combine deep learning and the reinforcement learning scheme to focus on the crucial dimensions of the signals.
\end{itemize}

\subsubsection{Results and Discussion} 
\label{sub:results_and_discussion_1}
First, we report the overall performance of all the compared algorithms.
From Table~\ref{tab:results_pamap2}, we can observe that the proposed approach (AVAE) outperforms all the algorithm-related and application-related state-of-the-art models, illustrating the effectiveness of the latent space in providing robust representations for easier semi-supervised classification. The advantage is demonstrated under all the supervision rates. 

\begin{figure*}[t]
\centering
     \begin{subfigure}[t]{0.25\textwidth}
        \centering
        \includegraphics[width=\textwidth]{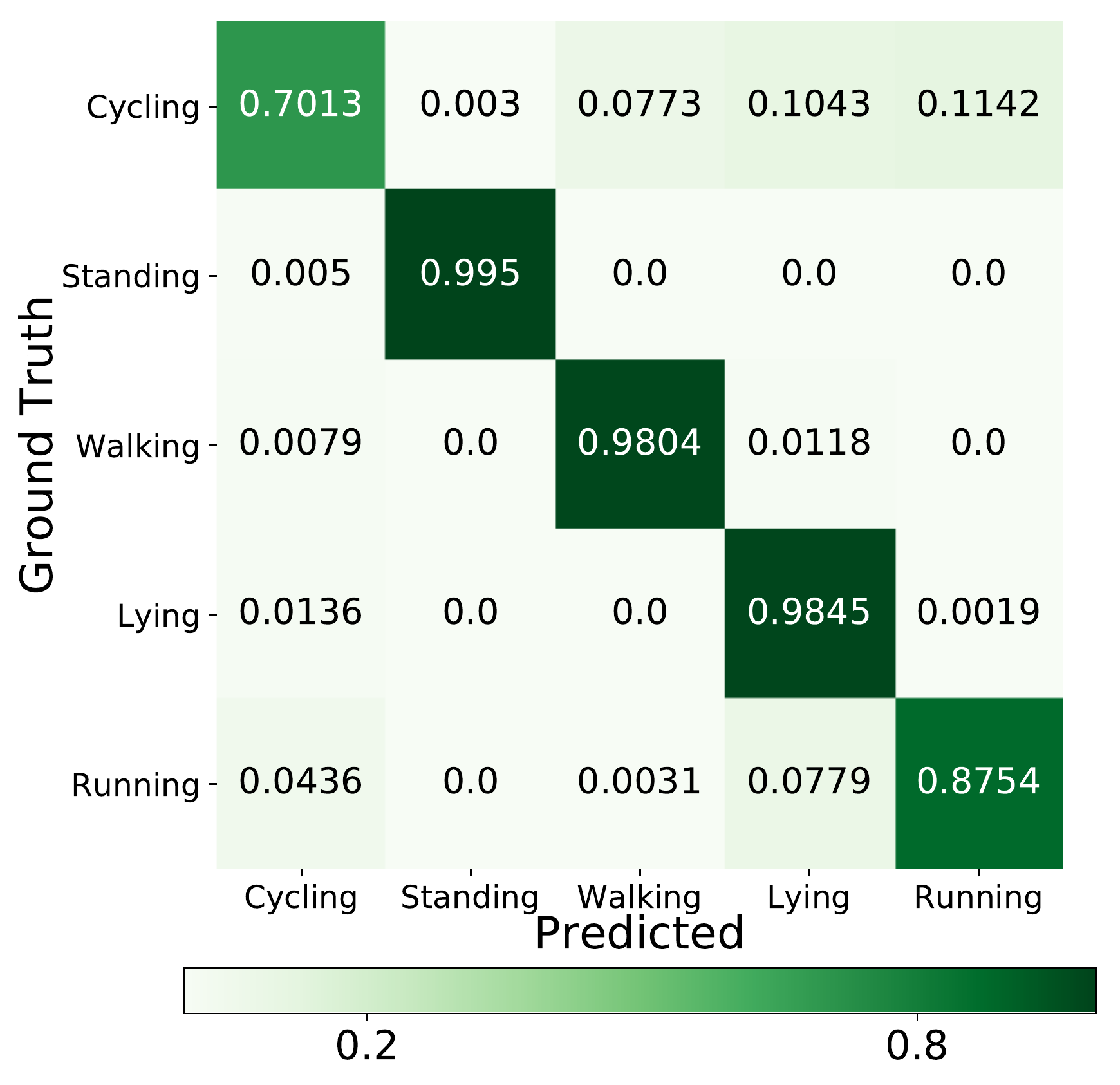}
        \caption{Confusion matrix of PAMAP2}
        \label{fig:pamap2_cm}
    \end{subfigure} 
	\begin{subfigure}[t]{0.25\textwidth}
        \centering
        \includegraphics[width=\textwidth]{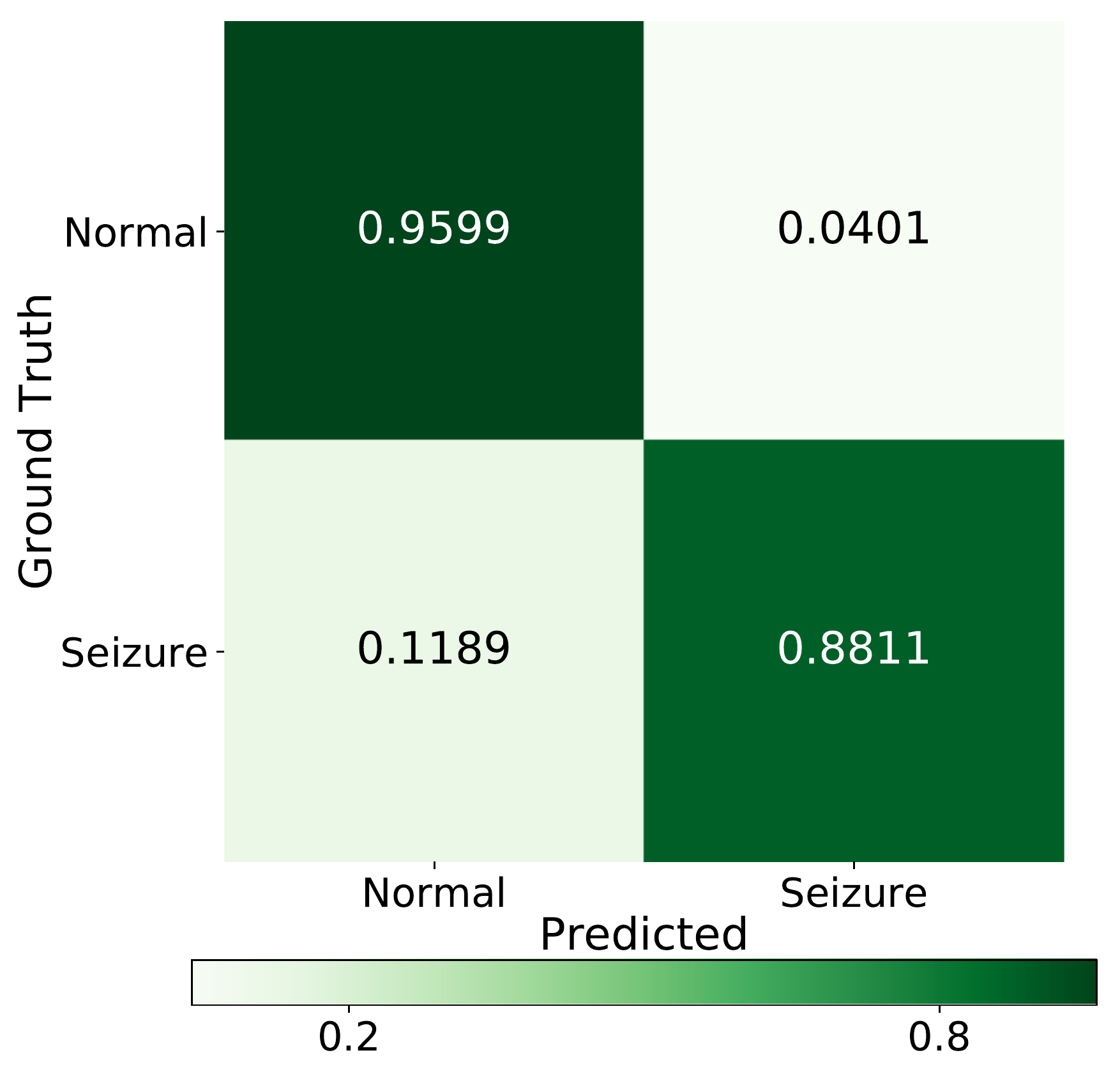}
        \caption{Confusion matrix of TUH}
        \label{fig:tuh_cm}    
    \end{subfigure}  
    \begin{subfigure}[t]{0.22\textwidth}
    \centering
    \includegraphics[width=\textwidth]{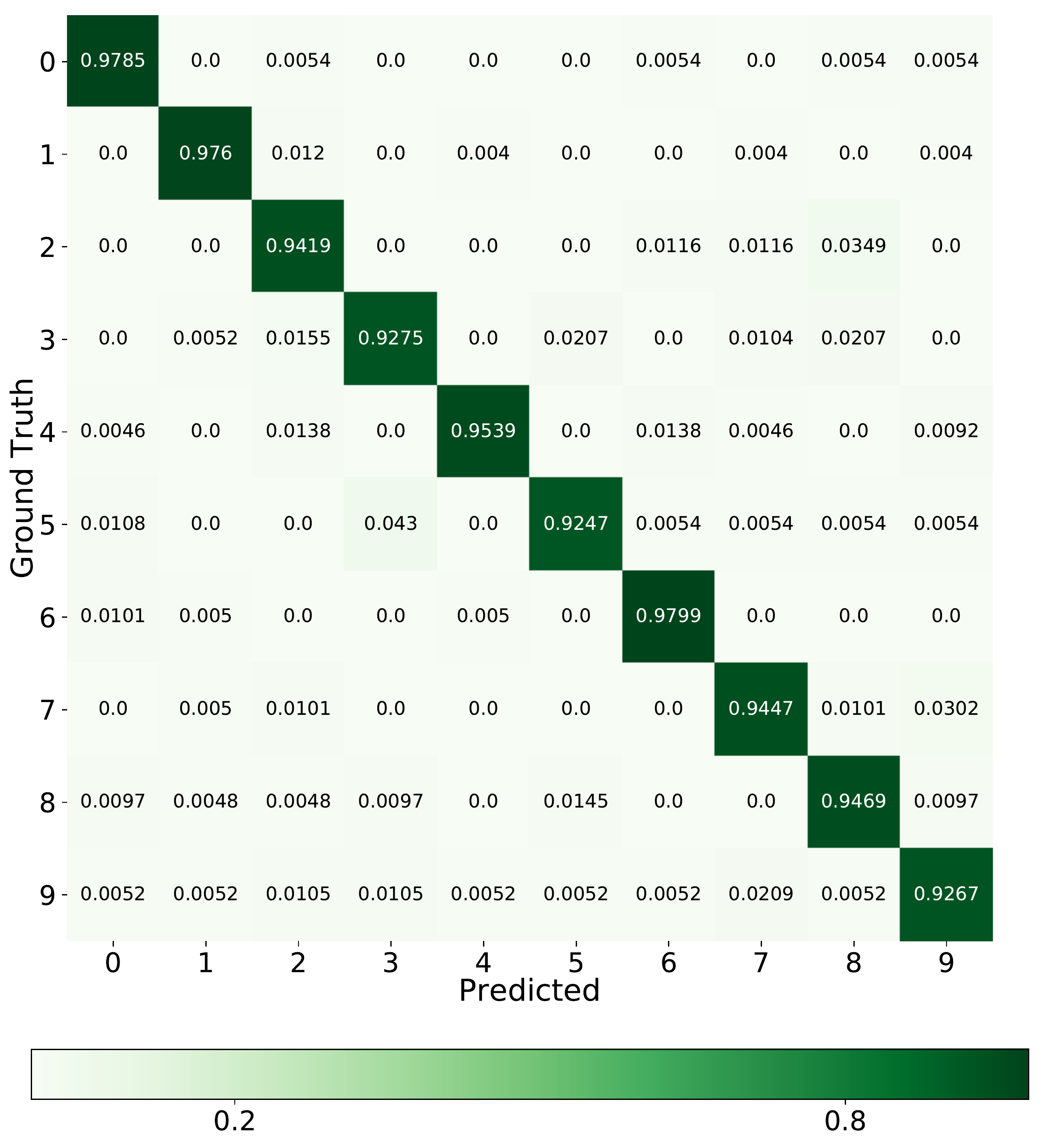}
    \caption{Confusion matrix of MNIST}
    \label{fig:mnist_cm}   
\end{subfigure}   
	\begin{subfigure}[t]{0.22\textwidth}
    \centering
    \includegraphics[width=\textwidth]{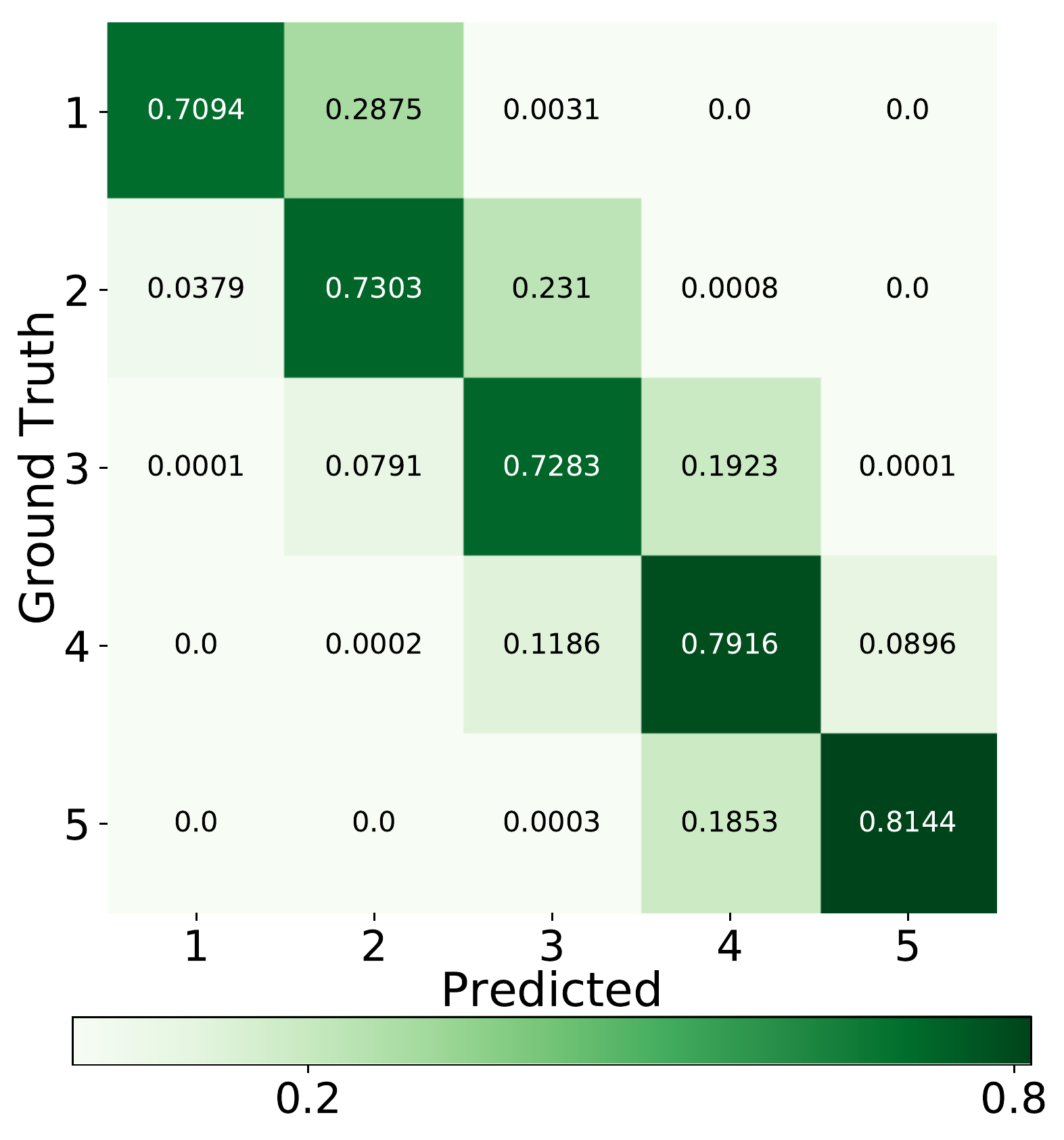}
    \caption{Confusion matrix of Yelp}
    \label{fig:yelp_cm}    
\end{subfigure} 
    \caption{Confusion matrix of PAMAP2, TUH, MNIST, and Yelp datasets.}
    \label{fig:cm}
 \vspace{-3mm}
\end{figure*}
\begin{figure*}[t]
\centering
     \begin{subfigure}[t]{0.24\textwidth}
        \centering
        \includegraphics[width=\textwidth]{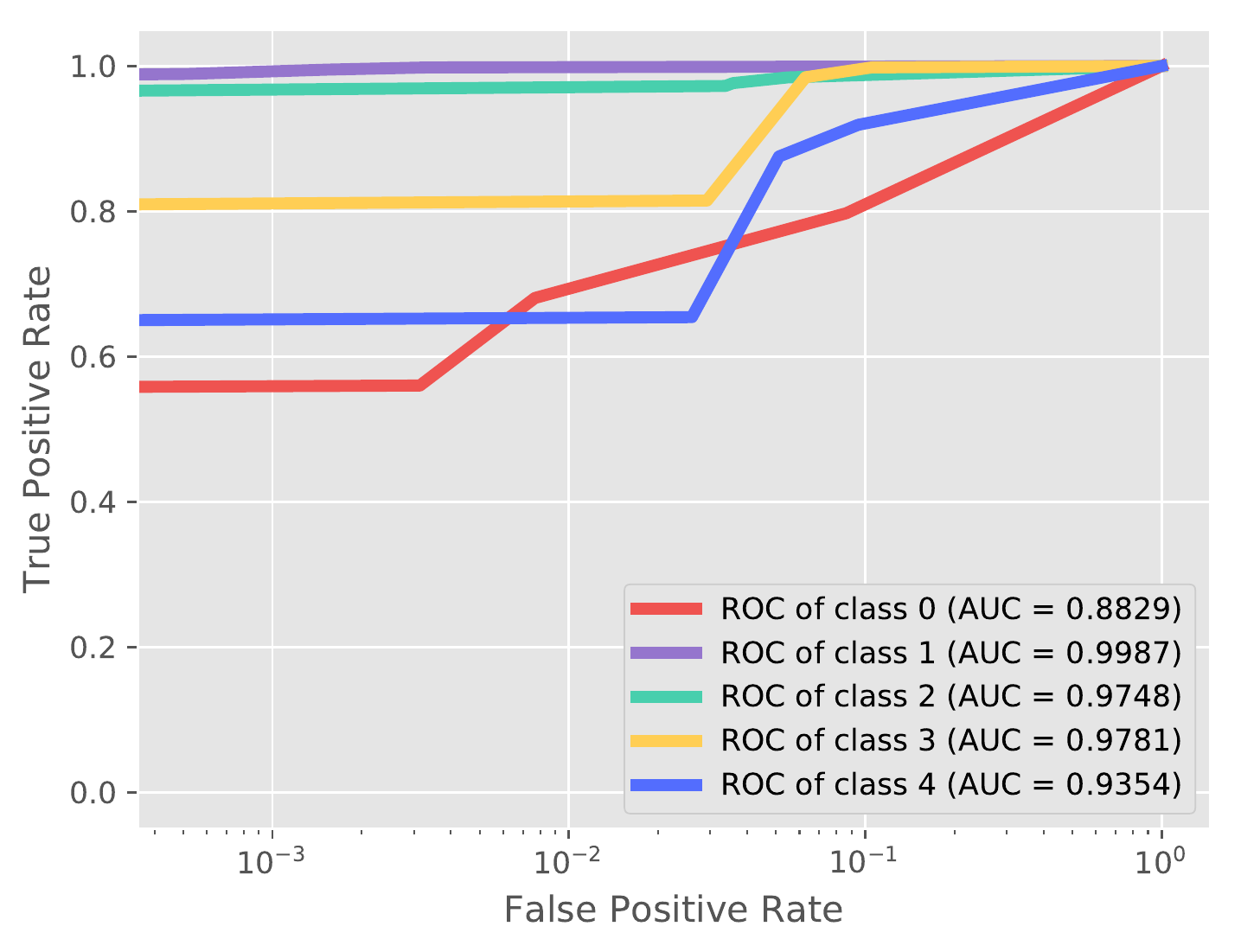}
        \caption{PAMAP2}
        \label{fig:pamap2_roc}
    \end{subfigure} 
	\begin{subfigure}[t]{0.24\textwidth}
        \centering
        \includegraphics[width=\textwidth]{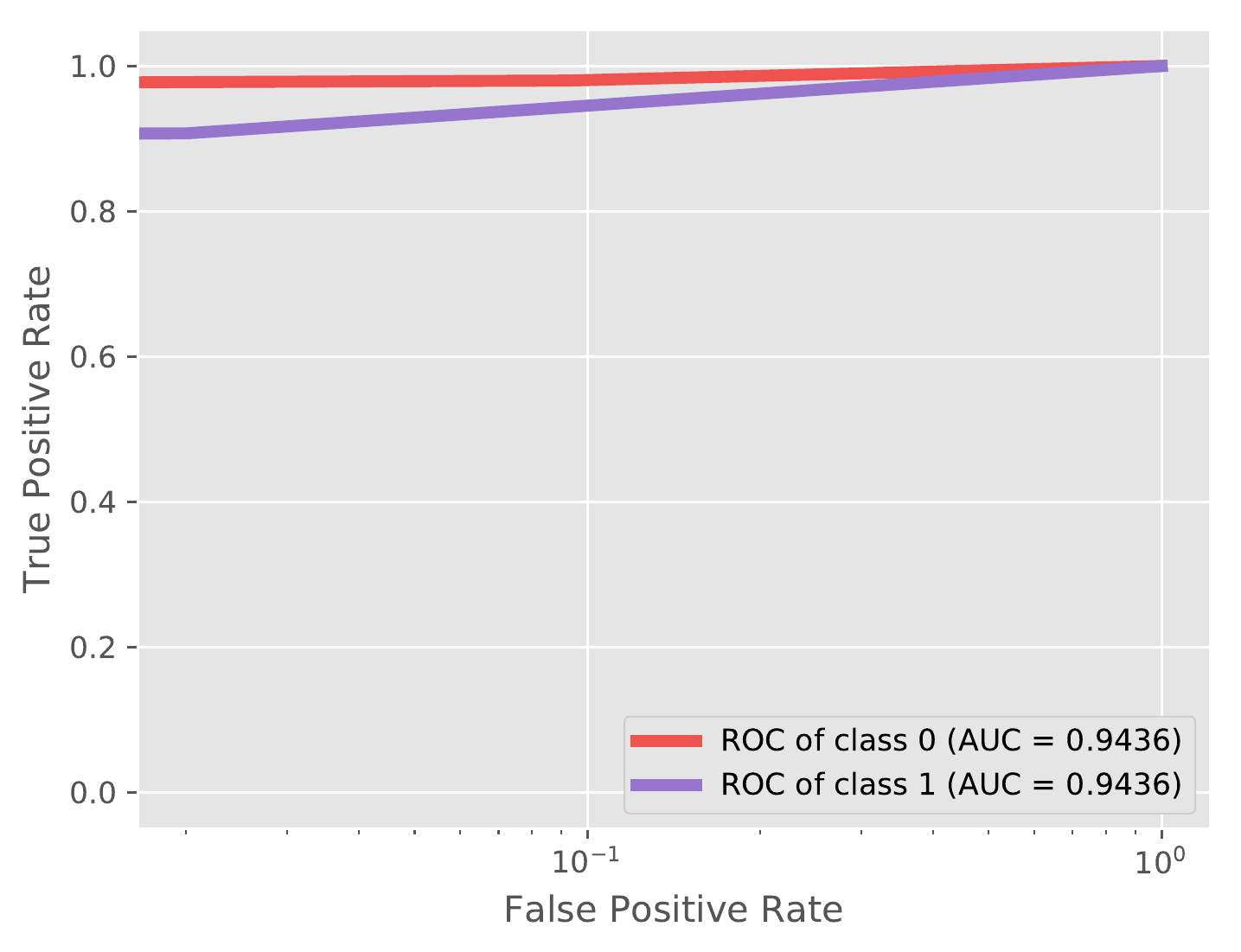}
        \caption{TUH}
        \label{fig:tuh_roc}    
    \end{subfigure}    
	\begin{subfigure}[t]{0.24\textwidth}
    \centering
    \includegraphics[width=\textwidth]{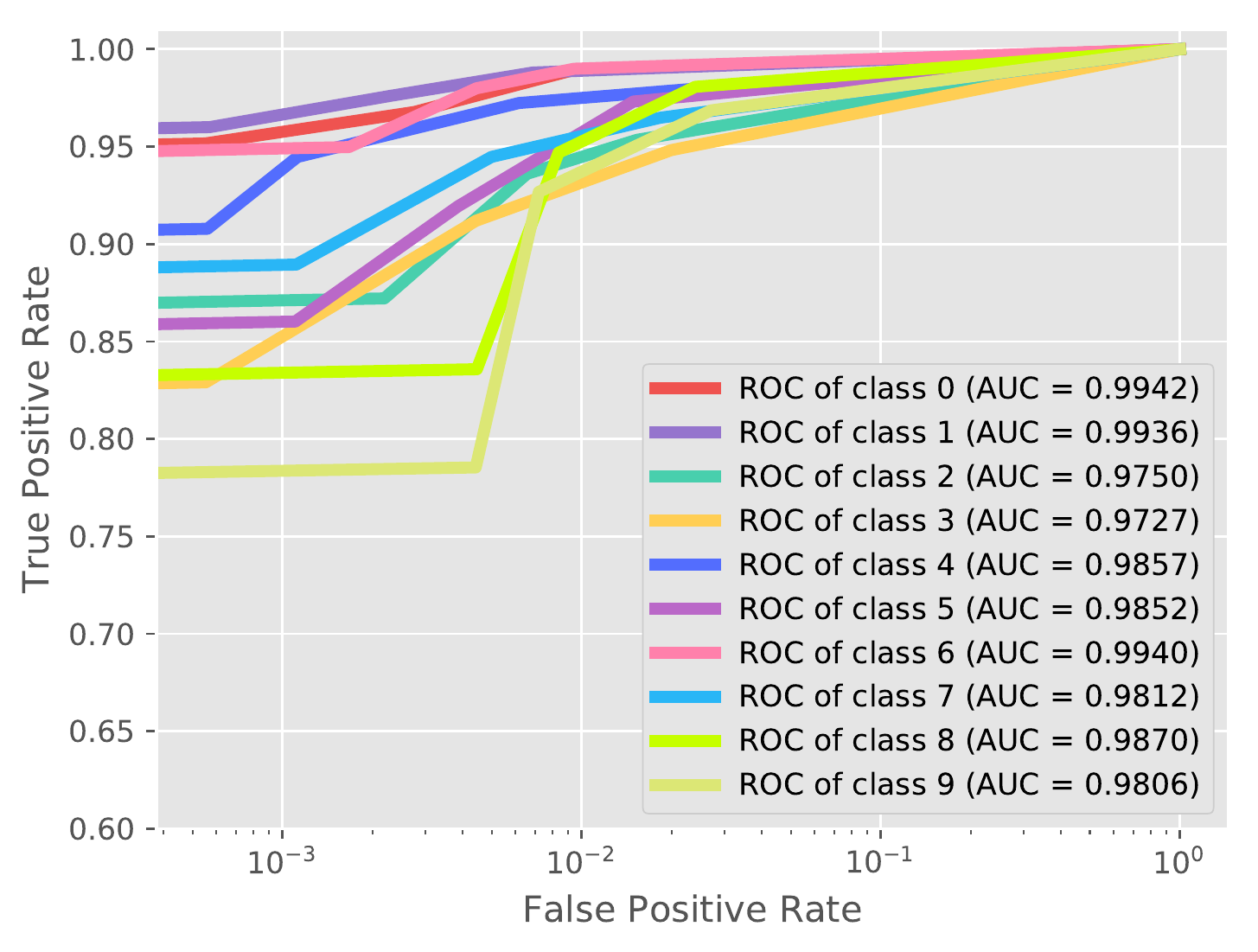}
    \caption{MNIST}
    \label{fig:mnist_roc}    
\end{subfigure} 
	\begin{subfigure}[t]{0.24\textwidth}
    \centering
    \includegraphics[width=\textwidth]{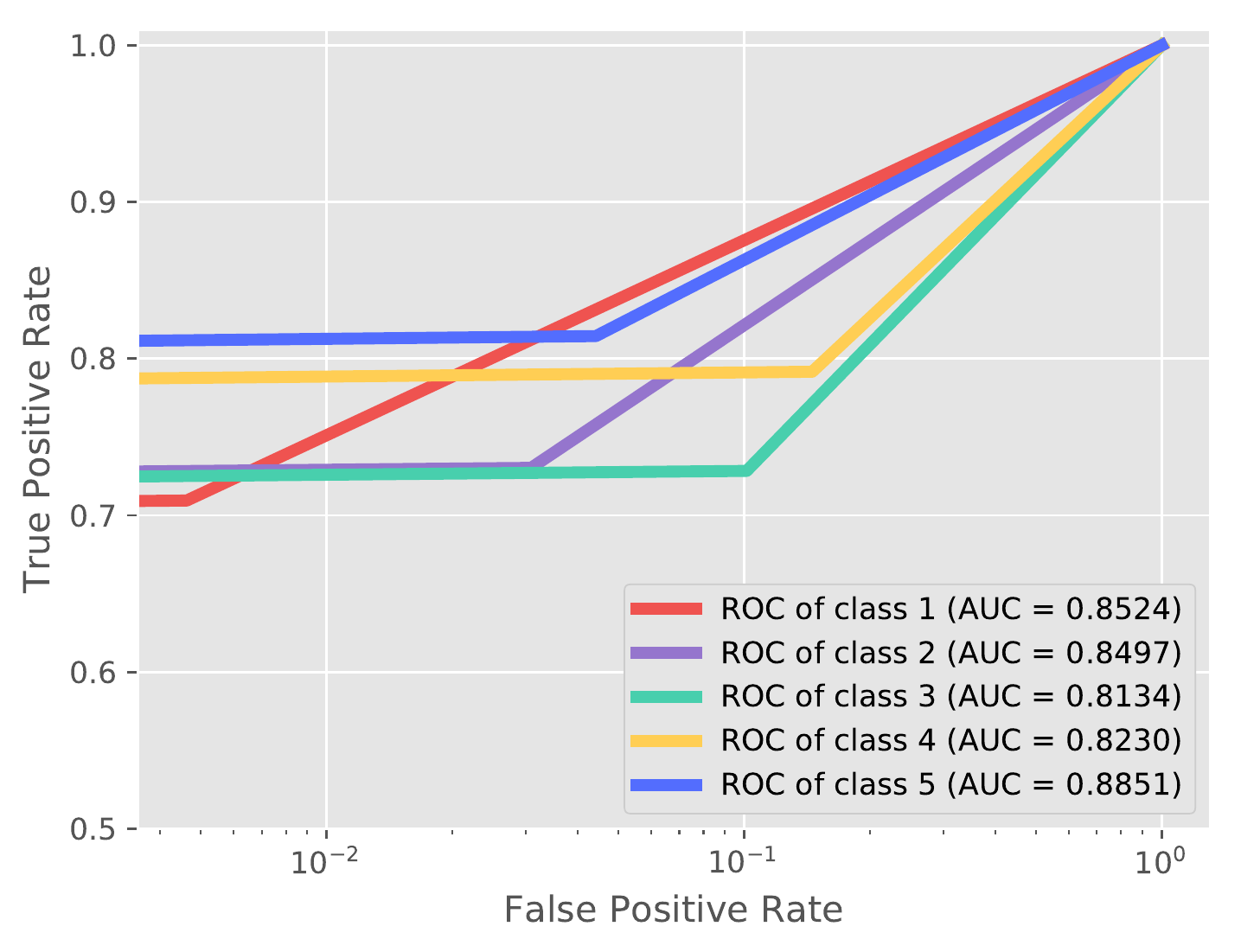}
    \caption{Yelp}
    \label{fig:yelp_roc}    
\end{subfigure}
    \caption{ROC curves of PAMAP2, TUH, MNIST, and Yelp datasets. The X-axis is in logarithmic scale.}
    \label{fig:roc}
     \vspace{-3mm}
\end{figure*}

In Table~\ref{tab:results_pamap2}, through the ablation study, it is observed that each component (especially GAN) contributes to the performance enhancement. Additionally, the proposed AVAE achieves a significant improvement which yields around $5\%$ and $3\%$ growth than the standard VAE and the VAE++ (under 60\% supervision rate), respectively. This observation demonstrates that the proposed latent layer $\bm{z}_I$ and the adversarial training (between the discriminator and VAE++) encourages the proposed model to learn and refine the informative latent code. Take 60\% supervision rate as an example, more details of the classification are shown in the confusion matrix (Figure~\ref{fig:pamap2_cm}) and ROC curves with AUC score (Figure~\ref{fig:pamap2_roc}).

\begin{figure*}[t]
    \centering
    \begin{subfigure}[t]{0.24\textwidth}
        \centering
        \includegraphics[width=\textwidth]{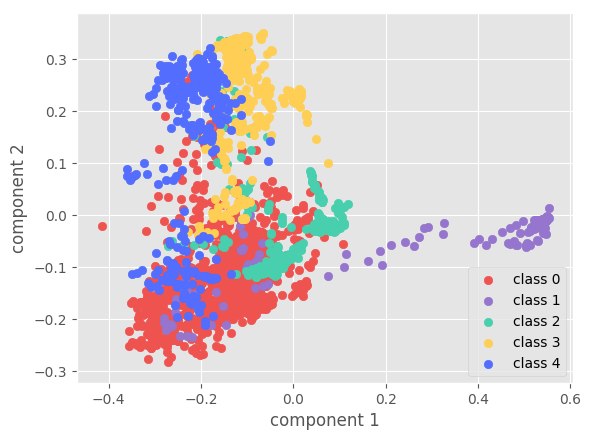}
        \caption{Raw (PAMAP2)}
    \end{subfigure}%
    \begin{subfigure}[t]{0.24\textwidth}
        \centering
        \includegraphics[width=\textwidth]{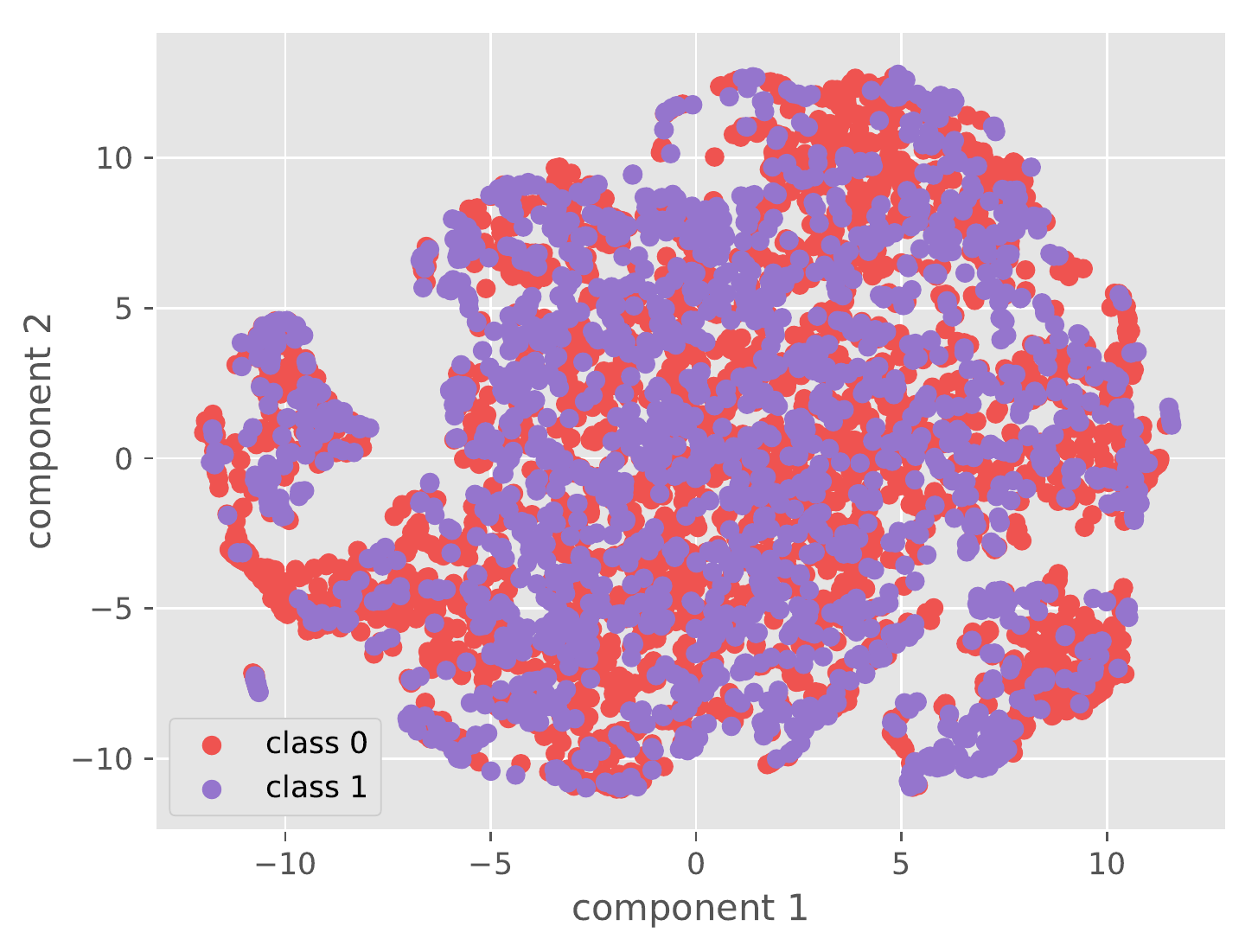}
        \caption{Raw (TUH)}
    \end{subfigure}%
    \begin{subfigure}[t]{0.24\textwidth}
        \centering
        \includegraphics[width=\textwidth]{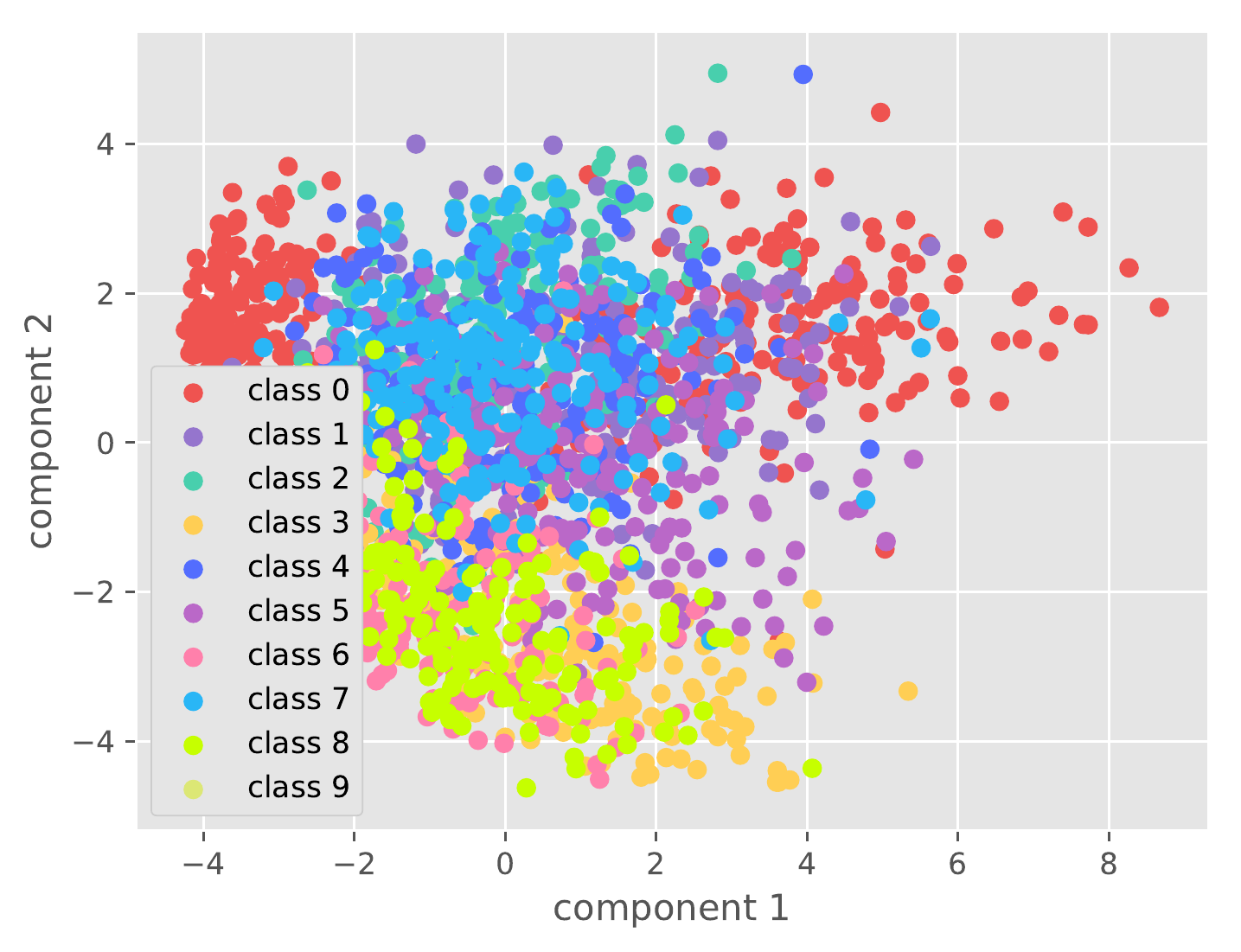}
        \caption{Raw (MNIST)}
    \end{subfigure}%
    \begin{subfigure}[t]{0.24\textwidth}
        \centering
        \includegraphics[width=\textwidth]{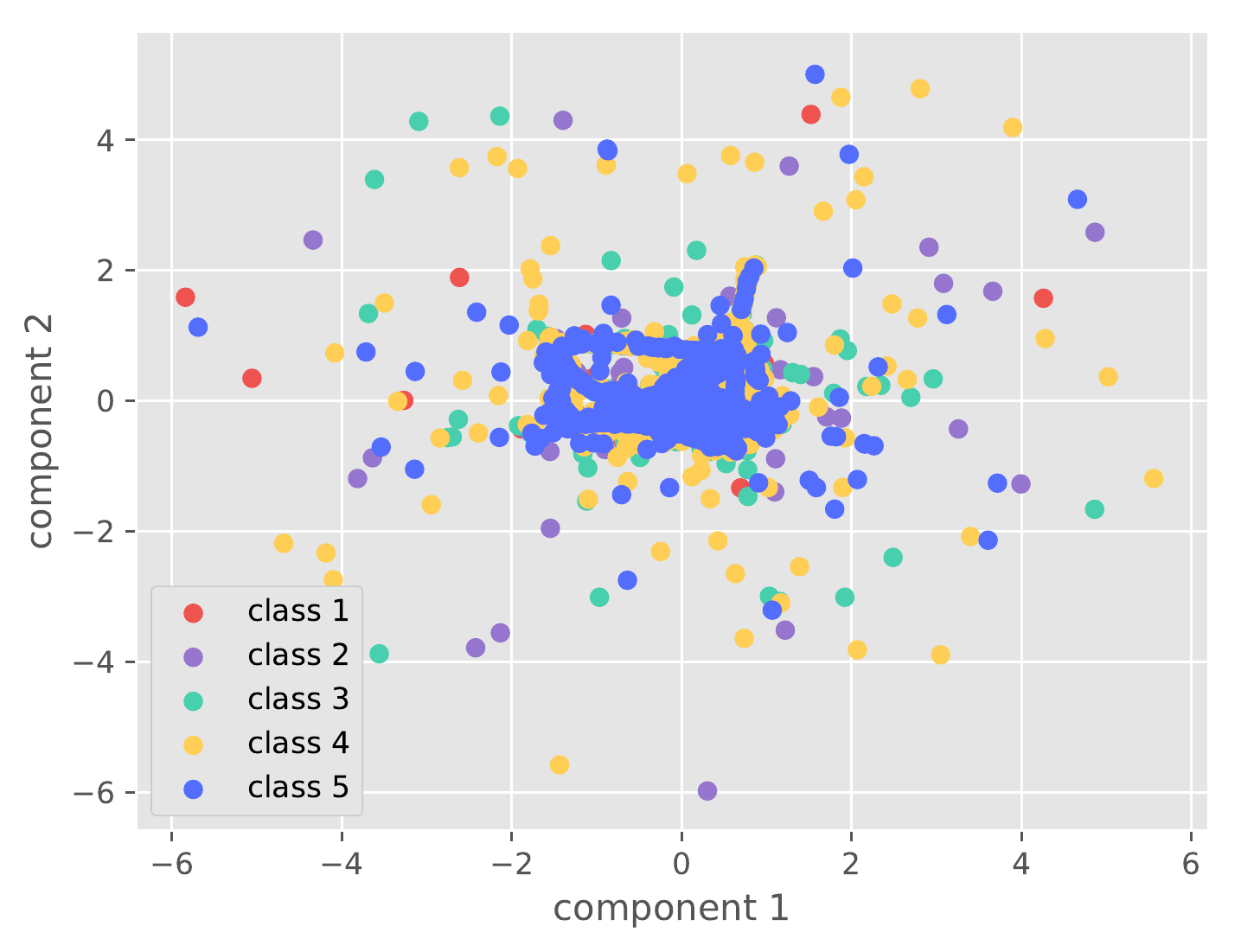}
        \caption{Raw (Yelp)}
    \end{subfigure}%

    \begin{subfigure}[t]{0.24\textwidth}
        \centering
        \includegraphics[width=\textwidth]{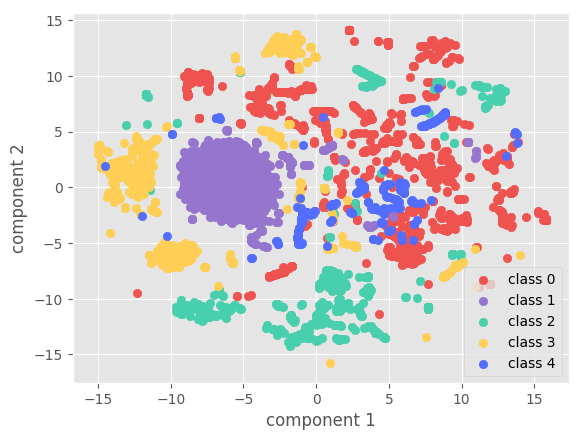}
        \caption{Learned feature (PAMAP2)}
    \end{subfigure}%
    \begin{subfigure}[t]{0.24\textwidth}
        \centering
        \includegraphics[width=\textwidth]{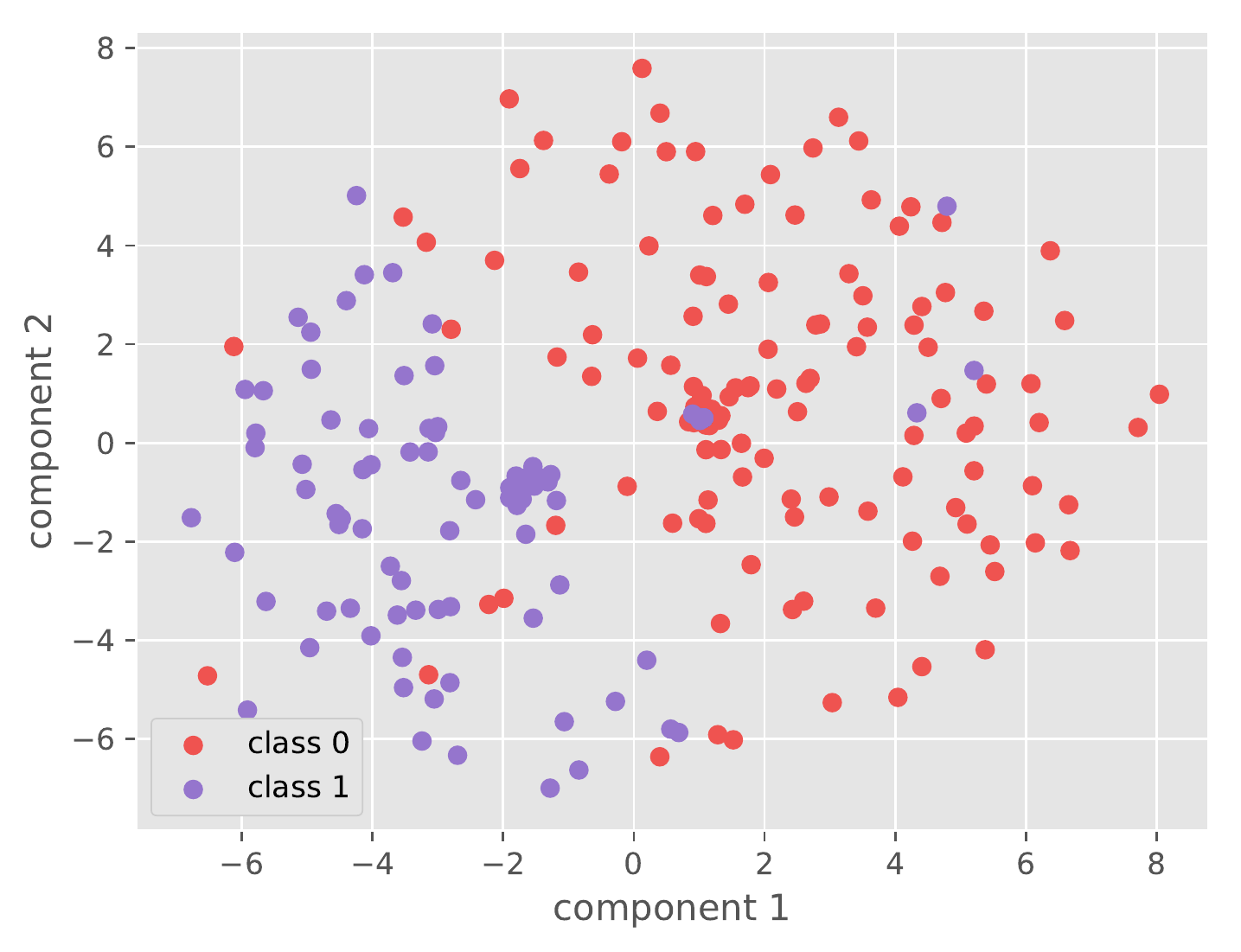}
        \caption{Learned feature (TUH)}
    \end{subfigure}%
    \begin{subfigure}[t]{0.24\textwidth}
        \centering
        \includegraphics[width=\textwidth]{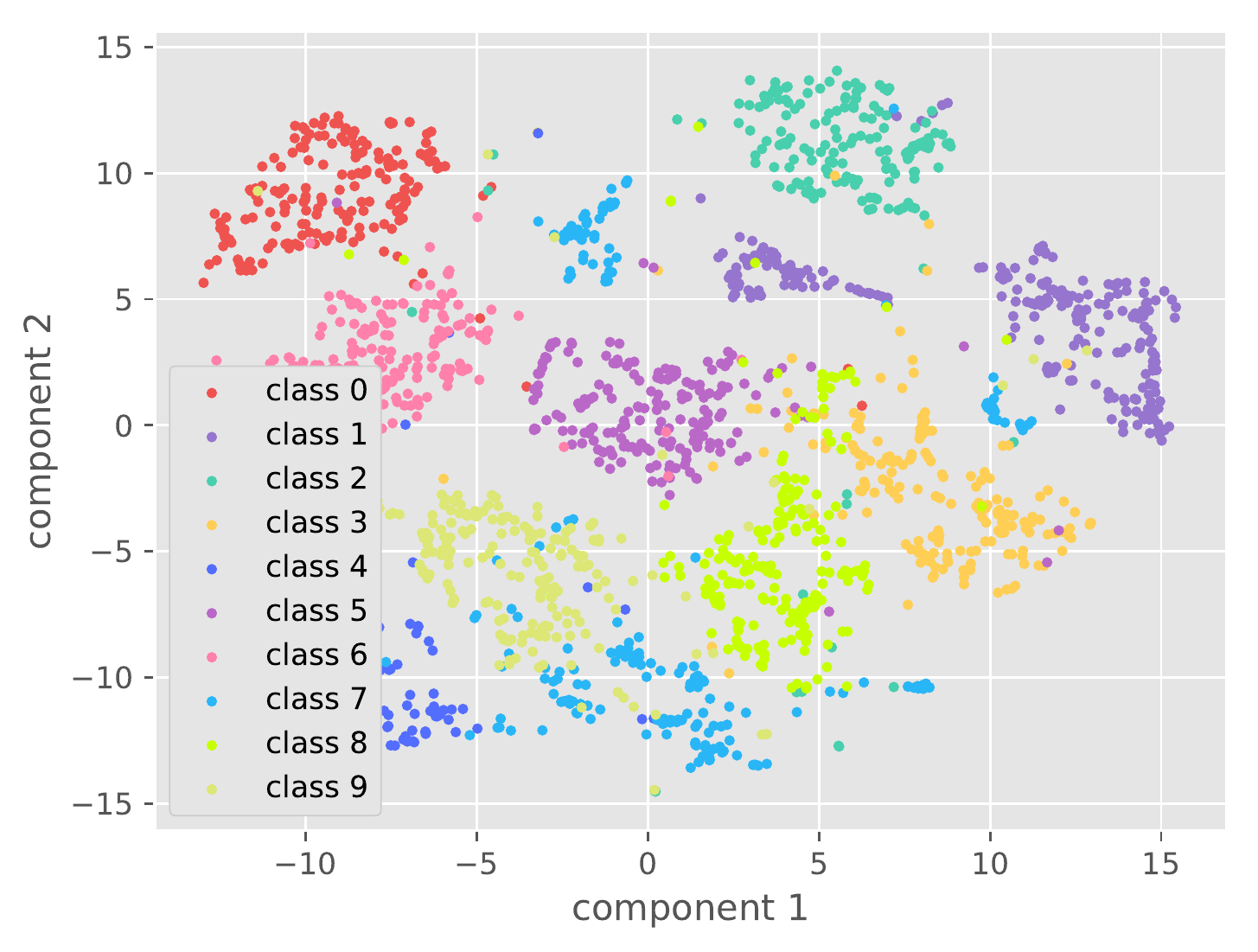}
        \caption{Learned feature (MNIST)}
    \end{subfigure}%
    \begin{subfigure}[t]{0.24\textwidth}
        \centering
        \includegraphics[width=\textwidth]{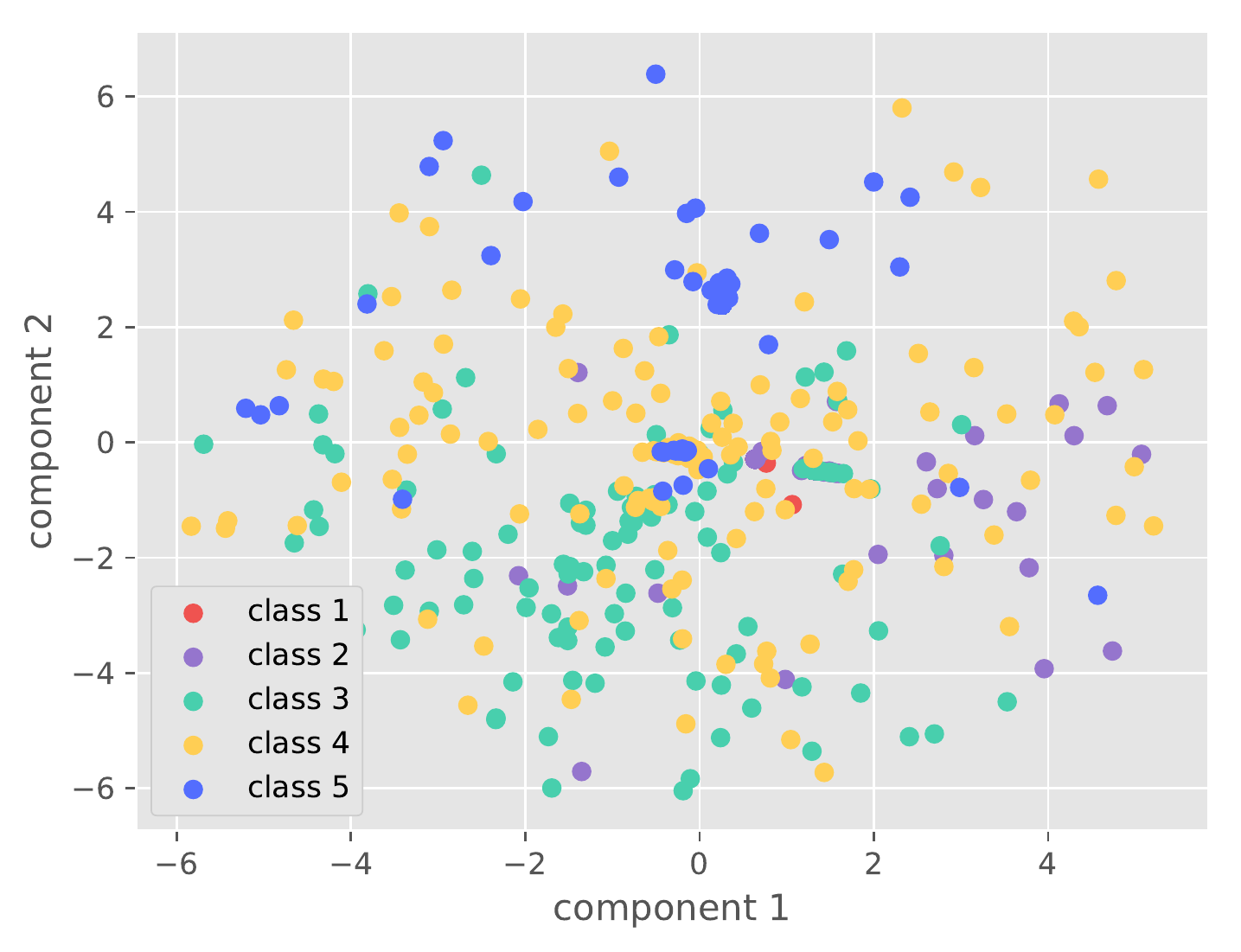}
        \caption{Learned feature (Yelp)}
    \end{subfigure}%
    \caption{Visualization comparison between raw data and the semi-supervised learned features
    }
    \label{fig:visual}
     \vspace{-3mm}
\end{figure*}

\begin{figure}[t]
    \centering
    \begin{subfigure}[t]{0.25\textwidth}
        \centering
        \includegraphics[width=\textwidth]{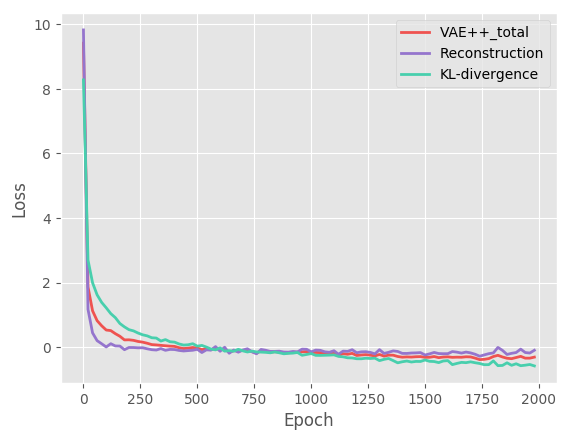}
        \caption{VAE++}
        \label{fig:loss_VAE}
    \end{subfigure}%
    \begin{subfigure}[t]{0.25\textwidth}
        \centering
        \includegraphics[width=\textwidth]{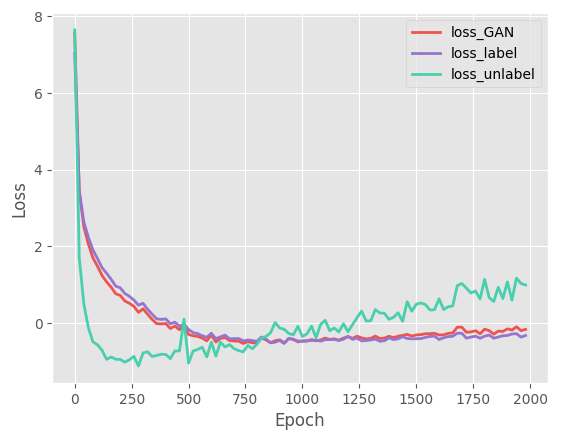}
        \caption{GAN}
        \label{fig:loss_GAN}
    \end{subfigure}%
    \caption{Convergence curve of the VAE++ and GAN }
    \label{fig:loss}
   \vspace{-5mm}
\end{figure}

\subsection{Neurological Diagnosis} 
\subsubsection{Experiment Setup} 
EEG signal collected in the unhealthy state differs significantly from the ones collected in the normal state \cite{adeli2007wavelet}. The epileptic seizure is a common brain disorder that affects about 1\% of the population and its octal state could be detected by the EEG analysis of the patient. In this application, we evaluate our framework with raw EEG data to diagnose the epileptic seizure of the patient.

We choose the benchmark dataset TUH \cite{obeid2016temple} for epileptic seizure diagnosis.
The TUH is a neurological seizure dataset of clinical EEG recordings associated with 22 channels from a 10/20 configuration. The sampling rate is set as 250 Hz. We select 12,000 samples from each of 18 subjects. Half of the samples are labelled as epileptic seizure state (labelled as 1) and the remaining samples are labelled as normal state (labelled as 0). The experiment and parameter settings are the same as the activity recognition applications.

\subsubsection{Baselines} 
The application-related state-of-the-art approaches in neurological diagnosis are listed here:
\begin{itemize}
	\item Ziyabari et al. \cite{ziyabari2017objective} adopt a hybrid deep learning architecture, including LSTM and stacked denoising Autoencoder, which integrates temporal and spatial context to detect the epileptic seizure.
	\item Harati et al. \cite{harati2015improved} demonstrate that a variant of the filter bank-based approach, coupled with first and second derivatives, provides a reduction in the overall error rate.
	\item Schimeister et al. \cite{schirrmeister2017deep} attempt to improve the performance of seizure detection by combining deep ConvNets with training strategies such as exponential linear units.
	\item Goodwin et al. \cite{goodwin2017deep} combine RNN with access to textual data in EEG reports in order to automatically extracting word- and report-level features and infer underspecified information from EHRs (electronic health records).
\end{itemize}

\subsubsection{Results and Discussion} 
From Table~\ref{tab:results_tuh}, we can observe that our approach outperforms all the competitive baselines on TUH dataset. For instance, under 60\% supervision level, the proposed approach achieves the highest accuracy of 95.21\% which claims around 4\% improvement over other methods. The corresponding confusion matrix (Figure~\ref{fig:tuh_cm}) and ROC curves (Figure~\ref{fig:tuh_roc}) infer that the normal state has higher accuracy than the seizure state. One possible reason is that the start and end stage of the seizure has similar symptoms with the normal state which may lead to misclassification.

\subsection{Image Classification} 
\subsubsection{Experiment Setup} 
To evaluate the representation learning ability in images, we test our approach on the benchmark dataset MNIST \footnote{http://yann.lecun.com/exdb/mnist/}. MNIST contains 60,000 handwritten digital images (50,000 for training and 10,000 for testing) with $28*28$ pixels. The labels of this dataset are from 0 to 9, corresponding to the 10 digits.

\subsubsection{Parameter Settings} 
\label{sub:parameter_settings_cv}
Images are more informative compared to other application scenarios. The encoder of AVAE is designed to be stacked by two convolutional layers. The first convolutional layer has 32 filters with shape $[3, 3]$, the stride size $[1, 1]$, 'SAME' padding, and ReLU activation function. The followed pooling layer has $[2,2]$ window size, $[2,2]$ stride, and 'SAME' padding. The second convolutional layer has 64 filters with $[5,5]$. The residual parameters of the second convolutional layer and the second pooling layer are the same with the former. 
Similarly, the decoder contains two de-convolutional layers with the same parameter settings. 
\subsubsection{Baselines} 
\label{sub:baselines}
We reproduce the following methods under different supervision rate for comparison:
\begin{itemize}
	\item Augustus \cite{odena2016semi} proposes a semi-supervised GAN (SGAN) by forcing the discriminator network to output class labels.
	\item Springenberg \cite{springenberg2015unsupervised} proposes CatGAN to modify the objective function taking into account the mutual information between observation and the prediction distribution.
	\item Weston et al. \cite{weston2012deep} apply kernel methods for a nonlinear semi-supervised embedding algorithm.
	\item Miyato et al. \cite{miyato2018virtual} propose a regularization method based on virtual adversarial loss: a new measure of local smoothness of the conditional label distribution given the inputs.
\end{itemize}

\subsubsection{Results and Discussion} 
As shown in Table~\ref{tab:results_mnist}, AVAE outperforms the counterparts with a slight gain with the same supervision level. The confusion matrix and ROC curves are reported in Figure~\ref{fig:mnist_cm} and Figure~\ref{fig:mnist_roc}. The results show that our approach is enabled to automatically learn the discriminative features by joint training the VAE++ and the semi-supervised GAN.

\subsection{Recommender System} 
\subsubsection{Experiment Setup} 
We apply our framework on recommender system scenarios, in particular, a restaurant rating prediction task based on the widely used Yelp dataset.

The Yelp Dataset\footnote{https://www.yelp.com/dataset} which includes 192,609 Businesses, 1,637,138 Users, and 6,685,900 Ratings. Each business has 13 attributes (like `near garage?', `have valet?') which can describe the quality and convenience of the business. Meanwhile, each business is rated by a series customers. The ratings range from 1 to 5, which can reflect the customers' satisfactory degree. 
Our recommender task considers a unseen business's attributes as input data and predict the possible ratings from the potential customers. If the rating is high enough, the new business will be recommended to the public. 

\subsubsection{Baselines} 
\label{sub:baselines}
We compare our approach with the state-of-the-art recommender system models which exploit the content information of items.
Since these methods are used to make rating predictions for each user-item pair, we select those users who have 200 and more ratings in the Yelp dataset, generating a set of 1,111 users. After collecting the predicted ratings for all user-item pairs, we take the average item ratings over the users, which are further rounded to serve as the predicted labels.

\begin{itemize}
    \item Pazzani et al. \cite{pazzani-2007-nn} summarizes basic content-based recommendation approaches, from which we select the cosine similarity-based nearest neighbour method as our fundamental baseline.

    \item Rendle \cite{rendle-2012-libfm} proposes the original implementation of factorization machine(FM) which is capable of incorporating item features with explicit feedbacks. We concatenate only the item indication vector and its feature after each user indication vector following the format in \cite{rendle-2012-libfm}. 

    \item He et al. \cite{he-2017-nfm} enhances the original FM using deep neural networks to learn high-order interactions between different item features.

    \item Chen et al. \cite{chen-2017-acf} applies feature- and item-level attention on item features, which is capable of emphasizing on the most important features.
\end{itemize}

\subsubsection{Results and Discussion} 
From Table~\ref{tab:results_yelp}, we can observe that our approach outperforms both the competitive semi-supervised algorithms and the content-based recommender system state-of-the-art methods. The rating prediction details can be found in Figure~\ref{fig:yelp_cm} and Figure~\ref{fig:yelp_roc}. The classification performance in recommender system is not good as in other applications. One possible reason is that the attributes data are very sparse. The experiment results illustrate that our approach is effective in recommender system scenarios.

\subsection{Further Analysis} 
\label{sub:further_analysis}
\subsubsection{Supervision Rate} 
\label{sub:supervision_rate}
We conduct extensive experiments to investigate the impact of supervision rate $\bm{\lambda}$. The supervision rate ranges from 20\% to 100\% with 20\% interval and each setting runs for at least three times with the average accuracy recorded. The overall performance varies with the supervision rate $\bm{\lambda}$. From Table~\ref{tab:results_tuh} to Table~\ref{tab:results_yelp}, it is noticed that the proposed model obtains competitive performance at each supervision level. 

\subsubsection{Visualization} 
\label{sub:visualization}
Figure~\ref{fig:visual} visualizes the raw data and the learned features on different datasets. The visualization comparison demonstrates the capability of our approach for distinguishable feature learning. 

\subsubsection{Convergence} 
\label{sub:convergence}

Take PAMAP2 as an example, Figure~\ref{fig:loss} presents the relationship between the loss function values and the epoch numbers. The VAE++ loss includes the reconstruction loss and the KL-divergence whilst the loss of the discriminator in GAN includes labelled loss and unlabelled loss (with weights 0.9 and 0.1, respectively). We can observe that the proposed method shows good convergence property as it stablizes in around 200 epochs.

\section{Conclusion} 
\label{sec:conclusion}
In this paper, we present an effective and robust semi-supervised latent representation framework, AVAE, by proposing a modified VAE model and integration with generative adversarial networks. The VAE++ and GAN share the same generator. In order to automatically learn the exclusive latent code, 
in the VAE++, 
we explore the latent code's posterior distribution and then stochastically generate a latent representation based on the posterior distribution. The discrepancy between the learned exclusive latent code and the generated latent representation is constrained by semi-supervised GAN.
The latent code of AVAE is finally served as the learned feature for classification. The proposed approach is evaluated on four real-world applications and the results demonstrate the effectiveness and robustness of our model.

The hyper-parameter tuning (not presented in this paper due to space limitation) in our model
is required for different datasets in various applications. One of our future scope is to propose a more generalized framework which is not sensitive to datasets. Moreover, our model still requires adequate labelled training samples for good performance. The lower supervision rate or unsupervised learning is another major goal in the future.

\vspace{-3mm}
\section{Acknowledgement} 
\label{sec:acknowledgements}
This research was partially supported by grant ONRG NICOP N62909-19-1-2009.

\bibliographystyle{ACM-Reference-Format}
\bibliography{kdd2019_cameready} 


\begin{thebibliography}{00}


\ifx \showCODEN    \undefined \def \showCODEN     #1{\unskip}     \fi
\ifx \showDOI      \undefined \def \showDOI       #1{{\tt DOI:}\penalty0{#1}\ }
  \fi
\ifx \showISBNx    \undefined \def \showISBNx     #1{\unskip}     \fi
\ifx \showISBNxiii \undefined \def \showISBNxiii  #1{\unskip}     \fi
\ifx \showISSN     \undefined \def \showISSN      #1{\unskip}     \fi
\ifx \showLCCN     \undefined \def \showLCCN      #1{\unskip}     \fi
\ifx \shownote     \undefined \def \shownote      #1{#1}          \fi
\ifx \showarticletitle \undefined \def \showarticletitle #1{#1}   \fi
\ifx \showURL      \undefined \def \showURL       #1{#1}          \fi
\providecommand\bibfield[2]{#2}
\providecommand\bibinfo[2]{#2}
\providecommand\natexlab[1]{#1}
\providecommand\showeprint[2][]{arXiv:#2}

\bibitem[\protect\citeauthoryear{Abbasnejad, Dick, and van~den
  Hengel}{Abbasnejad et~al\mbox{.}}{2017}]%
        {abbasnejad2017infinite}
\bibfield{author}{\bibinfo{person}{M~Ehsan Abbasnejad},
  \bibinfo{person}{Anthony Dick}, {and} \bibinfo{person}{Anton van~den
  Hengel}.} \bibinfo{year}{2017}\natexlab{}.
\newblock \showarticletitle{Infinite variational autoencoder for
  semi-supervised learning}. In \bibinfo{booktitle}{{\em 2017 IEEE Conference
  on Computer Vision and Pattern Recognition (CVPR)}}. IEEE,
  \bibinfo{pages}{781--790}.
\newblock


\bibitem[\protect\citeauthoryear{Adeli, Ghosh-Dastidar, and Dadmehr}{Adeli
  et~al\mbox{.}}{2007}]%
        {adeli2007wavelet}
\bibfield{author}{\bibinfo{person}{Hojjat Adeli}, \bibinfo{person}{Samanwoy
  Ghosh-Dastidar}, {and} \bibinfo{person}{Nahid Dadmehr}.}
  \bibinfo{year}{2007}\natexlab{}.
\newblock \showarticletitle{A wavelet-chaos methodology for analysis of EEGs
  and EEG subbands to detect seizure and epilepsy}.
\newblock \bibinfo{journal}{{\em IEEE Transactions on Biomedical
  Engineering\/}} \bibinfo{volume}{54}, \bibinfo{number}{2}
  (\bibinfo{year}{2007}), \bibinfo{pages}{205--211}.
\newblock


\bibitem[\protect\citeauthoryear{Bao, Chen, Wen, Li, and Hua}{Bao
  et~al\mbox{.}}{2017}]%
        {bao2017cvae}
\bibfield{author}{\bibinfo{person}{Jianmin Bao}, \bibinfo{person}{Dong Chen},
  \bibinfo{person}{Fang Wen}, \bibinfo{person}{Houqiang Li}, {and}
  \bibinfo{person}{Gang Hua}.} \bibinfo{year}{2017}\natexlab{}.
\newblock \showarticletitle{CVAE-GAN: fine-grained image generation through
  asymmetric training}.
\newblock \bibinfo{journal}{{\em CoRR, abs/1703.10155\/}}  \bibinfo{volume}{5}
  (\bibinfo{year}{2017}).
\newblock


\bibitem[\protect\citeauthoryear{Cao, Guo, Wu, Shen, and Tan}{Cao
  et~al\mbox{.}}{2018}]%
        {cao2018adversarial}
\bibfield{author}{\bibinfo{person}{Jiezhang Cao}, \bibinfo{person}{Yong Guo},
  \bibinfo{person}{Qingyao Wu}, \bibinfo{person}{Chunhua Shen}, {and}
  \bibinfo{person}{Mingkui Tan}.} \bibinfo{year}{2018}\natexlab{}.
\newblock \showarticletitle{Adversarial Learning with Local Coordinate Coding}.
\newblock \bibinfo{journal}{{\em The International Conference of Machine
  Learning (ICML)\/}} (\bibinfo{year}{2018}).
\newblock


\bibitem[\protect\citeauthoryear{Chen, Zhang, He, Nie, Liu, and Chua}{Chen
  et~al\mbox{.}}{2017}]%
        {chen-2017-acf}
\bibfield{author}{\bibinfo{person}{Jingyuan Chen}, \bibinfo{person}{Hanwang
  Zhang}, \bibinfo{person}{Xiangnan He}, \bibinfo{person}{Liqiang Nie},
  \bibinfo{person}{Wei Liu}, {and} \bibinfo{person}{Tat-Seng Chua}.}
  \bibinfo{year}{2017}\natexlab{}.
\newblock \showarticletitle{Attentive collaborative filtering: Multimedia
  recommendation with item-and component-level attention}. In
  \bibinfo{booktitle}{{\em SIGIR}}. ACM, \bibinfo{pages}{335--344}.
\newblock


\bibitem[\protect\citeauthoryear{Chen, Yao, Wang, Zhang, Gu, Yu, and Yang}{Chen
  et~al\mbox{.}}{2018}]%
        {chen2018interpretable}
\bibfield{author}{\bibinfo{person}{Kaixuan Chen}, \bibinfo{person}{Lina Yao},
  \bibinfo{person}{Xianzhi Wang}, \bibinfo{person}{Dalin Zhang},
  \bibinfo{person}{Tao Gu}, \bibinfo{person}{Zhiwen Yu}, {and}
  \bibinfo{person}{Zheng Yang}.} \bibinfo{year}{2018}\natexlab{}.
\newblock \showarticletitle{Interpretable Parallel Recurrent Neural Networks
  with Convolutional Attentions for Multi-Modality Activity Modeling}.
\newblock \bibinfo{journal}{{\em International Joint Conference on Neural
  Networks (IJCNN)\/}} (\bibinfo{year}{2018}).
\newblock


\bibitem[\protect\citeauthoryear{Chen, Duan, Houthooft, Schulman, Sutskever,
  and Abbeel}{Chen et~al\mbox{.}}{2016}]%
        {chen2016infogan}
\bibfield{author}{\bibinfo{person}{Xi Chen}, \bibinfo{person}{Yan Duan},
  \bibinfo{person}{Rein Houthooft}, \bibinfo{person}{John Schulman},
  \bibinfo{person}{Ilya Sutskever}, {and} \bibinfo{person}{Pieter Abbeel}.}
  \bibinfo{year}{2016}\natexlab{}.
\newblock \showarticletitle{Infogan: Interpretable representation learning by
  information maximizing generative adversarial nets}. In
  \bibinfo{booktitle}{{\em Advances in neural information processing systems}}.
  \bibinfo{pages}{2172--2180}.
\newblock


\bibitem[\protect\citeauthoryear{Fida, Bibbo, Bernabucci, Proto, Conforto, and
  Schmid}{Fida et~al\mbox{.}}{2015}]%
        {fida2015real}
\bibfield{author}{\bibinfo{person}{Benish Fida}, \bibinfo{person}{Daniele
  Bibbo}, \bibinfo{person}{Ivan Bernabucci}, \bibinfo{person}{Antonino Proto},
  \bibinfo{person}{Silvia Conforto}, {and} \bibinfo{person}{Maurizio Schmid}.}
  \bibinfo{year}{2015}\natexlab{}.
\newblock \showarticletitle{Real time event-based segmentation to classify
  locomotion activities through a single inertial sensor}. In
  \bibinfo{booktitle}{{\em MobiHealth}}. \bibinfo{pages}{104--107}.
\newblock


\bibitem[\protect\citeauthoryear{Ghasedi~Dizaji, Wang, and
  Huang}{Ghasedi~Dizaji et~al\mbox{.}}{2018}]%
        {ghasedi2018semi}
\bibfield{author}{\bibinfo{person}{Kamran Ghasedi~Dizaji},
  \bibinfo{person}{Xiaoqian Wang}, {and} \bibinfo{person}{Heng Huang}.}
  \bibinfo{year}{2018}\natexlab{}.
\newblock \showarticletitle{Semi-supervised generative adversarial network for
  gene expression inference}. In \bibinfo{booktitle}{{\em The 24th ACM SIGKDD
  International Conference on Knowledge Discovery \& Data Mining}}. ACM,
  \bibinfo{pages}{1435--1444}.
\newblock


\bibitem[\protect\citeauthoryear{Gong, Tao, Maybank, Liu, Kang, and Yang}{Gong
  et~al\mbox{.}}{2016}]%
        {gong2016multi}
\bibfield{author}{\bibinfo{person}{Chen Gong}, \bibinfo{person}{Dacheng Tao},
  \bibinfo{person}{Stephen~J Maybank}, \bibinfo{person}{Wei Liu},
  \bibinfo{person}{Guoliang Kang}, {and} \bibinfo{person}{Jie Yang}.}
  \bibinfo{year}{2016}\natexlab{}.
\newblock \showarticletitle{Multi-modal curriculum learning for semi-supervised
  image classification}.
\newblock \bibinfo{journal}{{\em IEEE Transactions on Image Processing\/}}
  \bibinfo{volume}{25}, \bibinfo{number}{7} (\bibinfo{year}{2016}),
  \bibinfo{pages}{3249--3260}.
\newblock


\bibitem[\protect\citeauthoryear{Goodwin and Harabagiu}{Goodwin and
  Harabagiu}{2017}]%
        {goodwin2017deep}
\bibfield{author}{\bibinfo{person}{Travis~R Goodwin} {and}
  \bibinfo{person}{Sanda~M Harabagiu}.} \bibinfo{year}{2017}\natexlab{}.
\newblock \showarticletitle{Deep Learning from EEG Reports for Inferring
  Underspecified Information}.
\newblock \bibinfo{journal}{{\em AMIA Summits on Translational Science
  Proceedings\/}}  \bibinfo{volume}{2017} (\bibinfo{year}{2017}),
  \bibinfo{pages}{112--121}.
\newblock


\bibitem[\protect\citeauthoryear{Guo, Chen, Peng, and Chen}{Guo
  et~al\mbox{.}}{2016}]%
        {guo2016wearable}
\bibfield{author}{\bibinfo{person}{Haodong Guo}, \bibinfo{person}{Ling Chen},
  \bibinfo{person}{Liangying Peng}, {and} \bibinfo{person}{Gencai Chen}.}
  \bibinfo{year}{2016}\natexlab{}.
\newblock \showarticletitle{Wearable sensor based multimodal human activity
  recognition exploiting the diversity of classifier ensemble}. In
  \bibinfo{booktitle}{{\em UbiComp}}. ACM, \bibinfo{pages}{1112--1123}.
\newblock


\bibitem[\protect\citeauthoryear{Harati, Golmohammadi, Lopez, Obeid, and
  Picone}{Harati et~al\mbox{.}}{2015}]%
        {harati2015improved}
\bibfield{author}{\bibinfo{person}{Amir Harati}, \bibinfo{person}{Meysam
  Golmohammadi}, \bibinfo{person}{Silvia Lopez}, \bibinfo{person}{Iyad Obeid},
  {and} \bibinfo{person}{Joseph Picone}.} \bibinfo{year}{2015}\natexlab{}.
\newblock \showarticletitle{Improved EEG event classification using
  differential energy}. In \bibinfo{booktitle}{{\em Signal Processing in
  Medicine and Biology Symposium (SPMB)}}. IEEE, \bibinfo{pages}{1--4}.
\newblock


\bibitem[\protect\citeauthoryear{He and Chua}{He and Chua}{2017}]%
        {he-2017-nfm}
\bibfield{author}{\bibinfo{person}{Xiangnan He} {and} \bibinfo{person}{Tat-Seng
  Chua}.} \bibinfo{year}{2017}\natexlab{}.
\newblock \showarticletitle{Neural factorization machines for sparse predictive
  analytics}. In \bibinfo{booktitle}{{\em Proceedings of the 40th International
  ACM SIGIR conference on Research and Development in Information Retrieval}}.
  ACM, \bibinfo{pages}{355--364}.
\newblock


\bibitem[\protect\citeauthoryear{Kingma, Mohamed, Rezende, and Welling}{Kingma
  et~al\mbox{.}}{2014}]%
        {kingma2014semi}
\bibfield{author}{\bibinfo{person}{Diederik~P Kingma}, \bibinfo{person}{Shakir
  Mohamed}, \bibinfo{person}{Danilo~Jimenez Rezende}, {and}
  \bibinfo{person}{Max Welling}.} \bibinfo{year}{2014}\natexlab{}.
\newblock \showarticletitle{Semi-supervised learning with deep generative
  models}. In \bibinfo{booktitle}{{\em Advances in Neural Information
  Processing Systems (NIPS)}}. \bibinfo{pages}{3581--3589}.
\newblock


\bibitem[\protect\citeauthoryear{Kingma and Welling}{Kingma and
  Welling}{2013}]%
        {kingma2013auto}
\bibfield{author}{\bibinfo{person}{Diederik~P Kingma} {and}
  \bibinfo{person}{Max Welling}.} \bibinfo{year}{2013}\natexlab{}.
\newblock \showarticletitle{Auto-encoding variational bayes}.
\newblock \bibinfo{journal}{{\em arXiv preprint arXiv:1312.6114\/}}
  (\bibinfo{year}{2013}).
\newblock


\bibitem[\protect\citeauthoryear{Krizhevsky, Sutskever, and Hinton}{Krizhevsky
  et~al\mbox{.}}{2012}]%
        {krizhevsky2012imagenet}
\bibfield{author}{\bibinfo{person}{Alex Krizhevsky}, \bibinfo{person}{Ilya
  Sutskever}, {and} \bibinfo{person}{Geoffrey~E Hinton}.}
  \bibinfo{year}{2012}\natexlab{}.
\newblock \showarticletitle{Imagenet classification with deep convolutional
  neural networks}. In \bibinfo{booktitle}{{\em Advances in neural information
  processing systems (NIPS)}}. \bibinfo{pages}{1097--1105}.
\newblock


\bibitem[\protect\citeauthoryear{Lara, P{\'e}rez, Labrador, and Posada}{Lara
  et~al\mbox{.}}{2012}]%
        {lara2012centinela}
\bibfield{author}{\bibinfo{person}{Oscar~D Lara}, \bibinfo{person}{Alfredo~J
  P{\'e}rez}, \bibinfo{person}{Miguel~A Labrador}, {and}
  \bibinfo{person}{Jos{\'e}~D Posada}.} \bibinfo{year}{2012}\natexlab{}.
\newblock \showarticletitle{Centinela: A human activity recognition system
  based on acceleration and vital sign data}.
\newblock \bibinfo{journal}{{\em Pervasive and mobile computing\/}}
  \bibinfo{volume}{8}, \bibinfo{number}{5} (\bibinfo{year}{2012}),
  \bibinfo{pages}{717--729}.
\newblock


\bibitem[\protect\citeauthoryear{Larsen, S{\o}nderby, Larochelle, and
  Winther}{Larsen et~al\mbox{.}}{2015}]%
        {larsen2015autoencoding}
\bibfield{author}{\bibinfo{person}{Anders Boesen~Lindbo Larsen},
  \bibinfo{person}{S{\o}ren~Kaae S{\o}nderby}, \bibinfo{person}{Hugo
  Larochelle}, {and} \bibinfo{person}{Ole Winther}.}
  \bibinfo{year}{2015}\natexlab{}.
\newblock \showarticletitle{Autoencoding beyond pixels using a learned
  similarity metric}.
\newblock \bibinfo{journal}{{\em arXiv preprint arXiv:1512.09300\/}}
  (\bibinfo{year}{2015}).
\newblock


\bibitem[\protect\citeauthoryear{Maal{\o}e, S{\o}nderby, S{\o}nderby, and
  Winther}{Maal{\o}e et~al\mbox{.}}{2016}]%
        {maaloe2016auxiliary}
\bibfield{author}{\bibinfo{person}{Lars Maal{\o}e},
  \bibinfo{person}{Casper~Kaae S{\o}nderby}, \bibinfo{person}{S{\o}ren~Kaae
  S{\o}nderby}, {and} \bibinfo{person}{Ole Winther}.}
  \bibinfo{year}{2016}\natexlab{}.
\newblock \showarticletitle{Auxiliary deep generative models}.
\newblock \bibinfo{journal}{{\em arXiv preprint:1602.05473\/}}
  (\bibinfo{year}{2016}).
\newblock


\bibitem[\protect\citeauthoryear{Makhzani, Shlens, Jaitly, Goodfellow, and
  Frey}{Makhzani et~al\mbox{.}}{2015}]%
        {makhzani2015adversarial}
\bibfield{author}{\bibinfo{person}{Alireza Makhzani}, \bibinfo{person}{Jonathon
  Shlens}, \bibinfo{person}{Navdeep Jaitly}, \bibinfo{person}{Ian Goodfellow},
  {and} \bibinfo{person}{Brendan Frey}.} \bibinfo{year}{2015}\natexlab{}.
\newblock \showarticletitle{Adversarial autoencoders}.
\newblock \bibinfo{journal}{{\em arXiv preprint arXiv:1511.05644\/}}
  (\bibinfo{year}{2015}).
\newblock


\bibitem[\protect\citeauthoryear{Mirza and Osindero}{Mirza and
  Osindero}{2014}]%
        {mirza2014conditional}
\bibfield{author}{\bibinfo{person}{Mehdi Mirza} {and} \bibinfo{person}{Simon
  Osindero}.} \bibinfo{year}{2014}\natexlab{}.
\newblock \showarticletitle{Conditional generative adversarial nets}.
\newblock \bibinfo{journal}{{\em arXiv preprint arXiv:1411.1784\/}}
  (\bibinfo{year}{2014}).
\newblock


\bibitem[\protect\citeauthoryear{Miyato, Maeda, Ishii, and Koyama}{Miyato
  et~al\mbox{.}}{2018}]%
        {miyato2018virtual}
\bibfield{author}{\bibinfo{person}{Takeru Miyato}, \bibinfo{person}{Shin-ichi
  Maeda}, \bibinfo{person}{Shin Ishii}, {and} \bibinfo{person}{Masanori
  Koyama}.} \bibinfo{year}{2018}\natexlab{}.
\newblock \showarticletitle{Virtual adversarial training: a regularization
  method for supervised and semi-supervised learning}.
\newblock \bibinfo{journal}{{\em IEEE transactions on pattern analysis and
  machine intelligence\/}} (\bibinfo{year}{2018}).
\newblock


\bibitem[\protect\citeauthoryear{Narayanaswamy, Paige, Van~de Meent, Desmaison,
  Goodman, Kohli, Wood, and Torr}{Narayanaswamy et~al\mbox{.}}{2017}]%
        {narayanaswamy2017learning}
\bibfield{author}{\bibinfo{person}{Siddharth Narayanaswamy},
  \bibinfo{person}{T~Brooks Paige}, \bibinfo{person}{Jan-Willem Van~de Meent},
  \bibinfo{person}{Alban Desmaison}, \bibinfo{person}{Noah Goodman},
  \bibinfo{person}{Pushmeet Kohli}, \bibinfo{person}{Frank Wood}, {and}
  \bibinfo{person}{Philip Torr}.} \bibinfo{year}{2017}\natexlab{}.
\newblock \showarticletitle{Learning disentangled representations with
  semi-supervised deep generative models}. In \bibinfo{booktitle}{{\em NIPS}}.
  \bibinfo{pages}{5925--5935}.
\newblock


\bibitem[\protect\citeauthoryear{Obeid and Picone}{Obeid and Picone}{2016}]%
        {obeid2016temple}
\bibfield{author}{\bibinfo{person}{Iyad Obeid} {and} \bibinfo{person}{Joseph
  Picone}.} \bibinfo{year}{2016}\natexlab{}.
\newblock \showarticletitle{The temple university hospital eeg data corpus}.
\newblock \bibinfo{journal}{{\em Frontiers in neuroscience\/}}
  \bibinfo{volume}{10} (\bibinfo{year}{2016}), \bibinfo{pages}{196}.
\newblock


\bibitem[\protect\citeauthoryear{Odena}{Odena}{2016}]%
        {odena2016semi}
\bibfield{author}{\bibinfo{person}{Augustus Odena}.}
  \bibinfo{year}{2016}\natexlab{}.
\newblock \showarticletitle{Semi-supervised learning with generative
  adversarial networks}.
\newblock \bibinfo{journal}{{\em arXiv preprint arXiv:1606.01583\/}}
  (\bibinfo{year}{2016}).
\newblock


\bibitem[\protect\citeauthoryear{Pazzani and Billsus}{Pazzani and
  Billsus}{2007}]%
        {pazzani-2007-nn}
\bibfield{author}{\bibinfo{person}{Michael~J Pazzani} {and}
  \bibinfo{person}{Daniel Billsus}.} \bibinfo{year}{2007}\natexlab{}.
\newblock \showarticletitle{Content-based recommendation systems}.
\newblock In \bibinfo{booktitle}{{\em The adaptive web}}.
  \bibinfo{publisher}{Springer}, \bibinfo{pages}{325--341}.
\newblock


\bibitem[\protect\citeauthoryear{Peng, Chen, Zhou, Li, Yang, and Zhang}{Peng
  et~al\mbox{.}}{2016}]%
        {peng2016immune}
\bibfield{author}{\bibinfo{person}{Lingxi Peng}, \bibinfo{person}{Wenbin Chen},
  \bibinfo{person}{Wubai Zhou}, \bibinfo{person}{Fufang Li},
  \bibinfo{person}{Jin Yang}, {and} \bibinfo{person}{Jiandong Zhang}.}
  \bibinfo{year}{2016}\natexlab{}.
\newblock \showarticletitle{An immune-inspired semi-supervised algorithm for
  breast cancer diagnosis}.
\newblock \bibinfo{journal}{{\em Computer methods and programs in
  biomedicine\/}}  \bibinfo{volume}{134} (\bibinfo{year}{2016}),
  \bibinfo{pages}{259--265}.
\newblock


\bibitem[\protect\citeauthoryear{Radford, Metz, and Chintala}{Radford
  et~al\mbox{.}}{2016}]%
        {radford2015unsupervised}
\bibfield{author}{\bibinfo{person}{Alec Radford}, \bibinfo{person}{Luke Metz},
  {and} \bibinfo{person}{Soumith Chintala}.} \bibinfo{year}{2016}\natexlab{}.
\newblock \showarticletitle{Unsupervised representation learning with deep
  convolutional generative adversarial networks}.
\newblock \bibinfo{journal}{{\em The International Conference on Learning
  Representations (ICLR)\/}} (\bibinfo{year}{2016}).
\newblock


\bibitem[\protect\citeauthoryear{Rendle}{Rendle}{2012}]%
        {rendle-2012-libfm}
\bibfield{author}{\bibinfo{person}{Steffen Rendle}.}
  \bibinfo{year}{2012}\natexlab{}.
\newblock \showarticletitle{Factorization machines with libfm}.
\newblock \bibinfo{journal}{{\em ACM Transactions on Intelligent Systems and
  Technology (TIST)\/}} \bibinfo{volume}{3}, \bibinfo{number}{3}
  (\bibinfo{year}{2012}), \bibinfo{pages}{57}.
\newblock


\bibitem[\protect\citeauthoryear{Salimans, Goodfellow, Zaremba, Cheung,
  Radford, and Chen}{Salimans et~al\mbox{.}}{2016}]%
        {salimans2016improved}
\bibfield{author}{\bibinfo{person}{Tim Salimans}, \bibinfo{person}{Ian
  Goodfellow}, \bibinfo{person}{Wojciech Zaremba}, \bibinfo{person}{Vicki
  Cheung}, \bibinfo{person}{Alec Radford}, {and} \bibinfo{person}{Xi Chen}.}
  \bibinfo{year}{2016}\natexlab{}.
\newblock \showarticletitle{Improved techniques for training gans}. In
  \bibinfo{booktitle}{{\em Advances in Neural Information Processing Systems
  (NIPS)}}. \bibinfo{pages}{2234--2242}.
\newblock


\bibitem[\protect\citeauthoryear{Schirrmeister, Gemein, Eggensperger, Hutter,
  and Ball}{Schirrmeister et~al\mbox{.}}{2017}]%
        {schirrmeister2017deep}
\bibfield{author}{\bibinfo{person}{Robin~Tibor Schirrmeister},
  \bibinfo{person}{Lukas Gemein}, \bibinfo{person}{Katharina Eggensperger},
  \bibinfo{person}{Frank Hutter}, {and} \bibinfo{person}{Tonio Ball}.}
  \bibinfo{year}{2017}\natexlab{}.
\newblock \showarticletitle{Deep learning with convolutional neural networks
  for decoding and visualization of EEG pathology}.
\newblock \bibinfo{journal}{{\em arXiv preprint:1708.08012\/}}
  (\bibinfo{year}{2017}).
\newblock


\bibitem[\protect\citeauthoryear{S{\o}nderby, Raiko, Maal{\o}e, S{\o}nderby,
  and Winther}{S{\o}nderby et~al\mbox{.}}{2016}]%
        {sonderby2016ladder}
\bibfield{author}{\bibinfo{person}{Casper~Kaae S{\o}nderby},
  \bibinfo{person}{Tapani Raiko}, \bibinfo{person}{Lars Maal{\o}e},
  \bibinfo{person}{S{\o}ren~Kaae S{\o}nderby}, {and} \bibinfo{person}{Ole
  Winther}.} \bibinfo{year}{2016}\natexlab{}.
\newblock \showarticletitle{Ladder variational autoencoders}. In
  \bibinfo{booktitle}{{\em Advances in neural information processing systems
  (NIPS)}}. \bibinfo{pages}{3738--3746}.
\newblock


\bibitem[\protect\citeauthoryear{Springenberg}{Springenberg}{2016}]%
        {springenberg2015unsupervised}
\bibfield{author}{\bibinfo{person}{Jost~Tobias Springenberg}.}
  \bibinfo{year}{2016}\natexlab{}.
\newblock \showarticletitle{Unsupervised and semi-supervised learning with
  categorical generative adversarial networks}.
\newblock \bibinfo{journal}{{\em The International Conference on Learning
  Representations (ICLR)\/}} (\bibinfo{year}{2016}).
\newblock


\bibitem[\protect\citeauthoryear{Walker, Doersch, Gupta, and Hebert}{Walker
  et~al\mbox{.}}{2016}]%
        {walker2016uncertain}
\bibfield{author}{\bibinfo{person}{Jacob Walker}, \bibinfo{person}{Carl
  Doersch}, \bibinfo{person}{Abhinav Gupta}, {and} \bibinfo{person}{Martial
  Hebert}.} \bibinfo{year}{2016}\natexlab{}.
\newblock \showarticletitle{An uncertain future: Forecasting from static images
  using variational autoencoders}. In \bibinfo{booktitle}{{\em European
  Conference on Computer Vision}}. Springer, \bibinfo{pages}{835--851}.
\newblock


\bibitem[\protect\citeauthoryear{Weston, Ratle, Mobahi, and Collobert}{Weston
  et~al\mbox{.}}{2012}]%
        {weston2012deep}
\bibfield{author}{\bibinfo{person}{Jason Weston},
  \bibinfo{person}{Fr{\'e}d{\'e}ric Ratle}, \bibinfo{person}{Hossein Mobahi},
  {and} \bibinfo{person}{Ronan Collobert}.} \bibinfo{year}{2012}\natexlab{}.
\newblock \showarticletitle{Deep learning via semi-supervised embedding}.
\newblock In \bibinfo{booktitle}{{\em Neural Networks: Tricks of the Trade}}.
  \bibinfo{publisher}{Springer}, \bibinfo{pages}{639--655}.
\newblock


\bibitem[\protect\citeauthoryear{Xu, Sun, Deng, and Tan}{Xu
  et~al\mbox{.}}{2017}]%
        {xu2017variational}
\bibfield{author}{\bibinfo{person}{Weidi Xu}, \bibinfo{person}{Haoze Sun},
  \bibinfo{person}{Chao Deng}, {and} \bibinfo{person}{Ying Tan}.}
  \bibinfo{year}{2017}\natexlab{}.
\newblock \showarticletitle{Variational Autoencoder for Semi-Supervised Text
  Classification.}. In \bibinfo{booktitle}{{\em AAAI}}.
  \bibinfo{pages}{3358--3364}.
\newblock


\bibitem[\protect\citeauthoryear{Yang, Bai, Zhang, Yuan, and Han}{Yang
  et~al\mbox{.}}{2017}]%
        {yang2017bridging}
\bibfield{author}{\bibinfo{person}{Carl Yang}, \bibinfo{person}{Lanxiao Bai},
  \bibinfo{person}{Chao Zhang}, \bibinfo{person}{Quan Yuan}, {and}
  \bibinfo{person}{Jiawei Han}.} \bibinfo{year}{2017}\natexlab{}.
\newblock \showarticletitle{Bridging collaborative filtering and
  semi-supervised learning: a neural approach for poi recommendation}. In
  \bibinfo{booktitle}{{\em Proceedings of the 23rd ACM SIGKDD International
  Conference on Knowledge Discovery and Data Mining}}. ACM,
  \bibinfo{pages}{1245--1254}.
\newblock


\bibitem[\protect\citeauthoryear{Yao, Nie, Sheng, Gu, Li, and Wang}{Yao
  et~al\mbox{.}}{2016}]%
        {yao2016learning}
\bibfield{author}{\bibinfo{person}{Lina Yao}, \bibinfo{person}{Feiping Nie},
  \bibinfo{person}{Quan~Z Sheng}, \bibinfo{person}{Tao Gu},
  \bibinfo{person}{Xue Li}, {and} \bibinfo{person}{Sen Wang}.}
  \bibinfo{year}{2016}\natexlab{}.
\newblock \showarticletitle{Learning from less for better: semi-supervised
  activity recognition via shared structure discovery}. In
  \bibinfo{booktitle}{{\em UbiComp}}. ACM, \bibinfo{pages}{13--24}.
\newblock


\bibitem[\protect\citeauthoryear{Zhang, Yao, Huang, Wang, Tan, Long, and
  Wang}{Zhang et~al\mbox{.}}{2018}]%
        {zhang2018multi}
\bibfield{author}{\bibinfo{person}{Xiang Zhang}, \bibinfo{person}{Lina Yao},
  \bibinfo{person}{Chaoran Huang}, \bibinfo{person}{Sen Wang},
  \bibinfo{person}{Mingkui Tan}, \bibinfo{person}{Guodong Long}, {and}
  \bibinfo{person}{Can Wang}.} \bibinfo{year}{2018}\natexlab{}.
\newblock \showarticletitle{Multi-modality Sensor Data Classification with
  Selective Attention}.
\newblock \bibinfo{journal}{{\em IJCAI\/}} (\bibinfo{year}{2018}).
\newblock


\bibitem[\protect\citeauthoryear{Ziyabari, Shah, Golmohammadi, Obeid, and
  Picone}{Ziyabari et~al\mbox{.}}{2017}]%
        {ziyabari2017objective}
\bibfield{author}{\bibinfo{person}{Saeedeh Ziyabari}, \bibinfo{person}{Vinit
  Shah}, \bibinfo{person}{Meysam Golmohammadi}, \bibinfo{person}{Iyad Obeid},
  {and} \bibinfo{person}{Joseph Picone}.} \bibinfo{year}{2017}\natexlab{}.
\newblock \showarticletitle{Objective evaluation metrics for automatic
  classification of EEG events}.
\newblock \bibinfo{journal}{{\em arXiv preprint arXiv:1712.10107\/}}
  (\bibinfo{year}{2017}).
\newblock


\end{thebibliography}

\end{document}